\documentclass[10pt,twocolumn,letterpaper]{article}

\usepackage{wacv}              

\usepackage{algorithm}
\usepackage{algorithm,algpseudocode}
\usepackage{algorithmicx}

\usepackage{graphicx}
\usepackage{amsmath}
\usepackage{amssymb}
\usepackage{booktabs}

\usepackage[utf8]{inputenc} 
\usepackage[T1]{fontenc}    
\usepackage{url}            
\usepackage{booktabs}       
\usepackage{amsfonts}       
\usepackage{nicefrac}       
\usepackage{microtype}      
\usepackage{xcolor}         

\usepackage{color}
\usepackage{amsmath}
\usepackage{amssymb}
\usepackage{mathtools}
\usepackage{arydshln}
\usepackage{amsthm}
\usepackage{bm}
\usepackage{enumitem}
\usepackage{multirow}

\usepackage[pagebackref,breaklinks,colorlinks]{hyperref}

\usepackage[capitalize,noabbrev]{cleveref}

\usepackage[capitalize]{cleveref}
\crefname{section}{Sec.}{Secs.}
\Crefname{section}{Section}{Sections}
\Crefname{table}{Table}{Tables}
\crefname{table}{Tab.}{Tabs.}

\newcommand{\bx}{\mathbf{x}}

\newcommand{\xsrc}{\mathbf{x}^{\mathrm{src}}}
\newcommand{\xsrct}{\mathbf{x}^{\mathrm{src}}_t}
\newcommand{\bxtgt}{\bar{\mathbf{x}}^{\mathrm{tgt}}}
\newcommand{\bxtgtt}{\bar{\mathbf{x}}^{\mathrm{tgt}}_t}
\newcommand{\bxsrc}{\bar{\mathbf{x}}^{\mathrm{src}}}
\newcommand{\xtgt}{\mathbf{x}^{\mathrm{tgt}}}
\newcommand{\xtgtt}{\mathbf{x}^{\mathrm{tgt}}_t}

\newcommand{\ysrc}{\mathbf{y}^{\mathrm{src}}}
\newcommand{\y}{\mathbf{y}}
\newcommand{\ytgt}{\mathbf{y}^{\mathrm{tgt}}}
\newcommand{\psrc}{p^{\mathrm{src}}}
\newcommand{\ptgt}{p^{\mathrm{tgt}}}

\newcommand{\bred}[1]{\textbf{\textcolor{red}{#1}}}
\newcommand{\black}[1]{\textbf{\textcolor{black}{#1}}}

\def\wacvPaperID{1977} 
\def\confName{WACV}
\def\confYear{2025}

\begin{document}

\title{Diffusion-Based Conditional Image Editing through\\Optimized Inference with Guidance}

\author{
Hyunsoo Lee\textsuperscript{\normalfont 1} \hspace{10mm} Minsoo Kang\textsuperscript{\normalfont 1} \hspace{10mm} Bohyung Han\textsuperscript{\normalfont 1,2} \\ 
   \textsuperscript{1}ECE \&  \textsuperscript{2}IPAI, Seoul National University \hspace{10mm} \\
   {\tt\small \{philip21,\,kminsoo,\,bhhan\}@snu.ac.kr}
}

\maketitle



\begin{abstract}
We present a simple but effective training-free approach for text-driven image-to-image translation based on a pretrained text-to-image diffusion model.
Our goal is to generate an image that aligns with the target task while preserving the structure and background of a source image.
To this end, we derive the representation guidance with a combination of two objectives: maximizing the similarity to the target prompt based on the CLIP score and minimizing the structural distance to the source latent variable.
This guidance improves the fidelity of the generated target image to the given target prompt while maintaining the structure integrity of the source image.
To incorporate the representation guidance component, we optimize the target latent variable of diffusion model's reverse process with the guidance.
Experimental results demonstrate that our method achieves outstanding image-to-image translation performance on various tasks when combined with the pretrained Stable Diffusion model.
\end{abstract} 



\section{Introduction}
\label{sec:introduction}

Diffusion-based text-to-image generation models~\cite{rombach2022high, ramesh2022hierarchical, kawar2023imagic} have shown superior performance in generating high-quality images.
These models have also been adapted for text-driven image editing tasks, aiming to translate a given source image into a target domain while preserving its overall structure and background.
However, text-driven image-to-image translation remains a challenging task, as it requires selectively modifying specific parts of the image while maintaining the integrity of the background and structure.
Moreover, achieving precise control over fine details further adds to the complexity of this task.

To address these tasks, many text-driven image editing methods~\cite{kim2022diffusionclip, valevski2022unitune, kawar2023imagic, brooks2023instructpix2pix} rely on additional fine-tuning of pretrained diffusion models.
While these methods yield promising results, they are impractical due to the substantial computational and memory overhead required for the fine-tuning process.
In contrast, training-free algorithms~\cite{hertz2022prompt, tumanyan2023plug, parmar2023zero, couairon2022diffedit, mokady2023null} introduce unique inference techniques for diffusion models, achieving their objectives without fine-tuning.
However, despite their efficiency, these approaches often struggle to preserve the structure of the source image and tend to produce blurry results.

We propose a simple yet effective training-free image-to-image translation method built upon pretrained diffusion models.
Our approach incorporates representation guidance based on a triplet loss to enhance fidelity to the target task.
Specifically, we modify the reverse process of the diffusion model by integrating the representation guidance, which comprises two key components:
(a) a CLIP~\cite{radford2021learning}-based objective that ensures the sampled target latent aligns semantically with the given target prompt, and
(b) a structural constraint that enforces similarity between the source and target images' structural information, as extracted from the feature maps of the pretrained diffusion model.
Experimental results demonstrate that the proposed reverse process effectively preserves the background and structure of image for both local and global editing tasks.
The main contributions of our work are summarized below:
\begin{itemize}[label=$\bullet$]
	\item
	We propose a novel representation guidance mechanism motivated by metric learning, leveraging features from the pretrained CLIP and Stable Diffusion for text-driven image-to-image translation.
    \item Our approach modifies the denoising process of the pretrained Stable Diffusion model without any additional training procedure.
    \item Experimental results on various image-to-image translation tasks using both real and synthetic images verify the outstanding performance of our approach compared to existing methods.
\end{itemize}
%


\section{Related Work}
\label{sec:related_work}

\subsection{Text-to-image diffusion models}
\label{sec:related_T2I}
Existing methods~\cite{rombach2022high, saharia2022photorealistic, ramesh2022hierarchical} based on diffusion models have demonstrated remarkable performance in text-to-image generation tasks.
For instance, Stable Diffusion~\cite{rombach2022high} leverages pretrained autoencoders~\cite{kingma2014autoencoding} to project a given image onto a low-dimensional latent space and estimates its distribution within this manifold, rather than modeling the raw data distribution directly.
Imagen~\cite{saharia2022photorealistic} employs a large pretrained text encoder to generate text embeddings, which are then used to condition the diffusion model for image synthesis.
Similarly, DALL$\cdot$E 2~\cite{ramesh2022hierarchical} predicts the CLIP image embedding from a text caption and synthesizes an image conditioned on both the estimated CLIP embedding and the textual input.

\subsection{Diffusion-based image manipulation methods}
\label{sec:related_I2I}

Text-driven image manipulation aims to preserve the structure and background of the source image while selectively editing the image to align with the target prompt.
Several works~\cite{parmar2023zero, dong2023prompt, couairon2022diffedit, hertz2022prompt, tumanyan2023plug, epstein2023diffusion, lee2023conditional, lee2025diffusion} employ publicly available pretrained text-to-image diffusion models, such as Stable Diffusion~\cite{rombach2022high}, to deal with the image editing tasks.
For example, DiffEdit~\cite{couairon2022diffedit} adaptively interpolates between the source and target latents at each time step, guided by an estimated object mask.
Prompt-to-Prompt~\cite{hertz2022prompt} substitutes the self-attention and cross-attention maps of the source latents for those retrieved from the target latents.
Plug-and-Play~\cite{tumanyan2023plug} injects the self-attention and intermediate feature maps obtained from the source latents into the target image generation process. 
Pix2Pix-Zero~\cite{parmar2023zero}, on the other hand, optimizes the target latent to align with cross-attention maps extracted from the pretrained diffusion model, based on both the source and target latents.
Null-text inversion~\cite{mokady2023null} first inverts a source image using an optimization-based pivotal tuning procedure to achieve precise reconstruction and then applies the Prompt-to-Prompt technique to generate the target image based on the inverted image.
MasaCtrl~\cite{cao2023masactrl} modifies traditional self-attention mechanisms in diffusion models, allowing the model to utilize the local features extracted from the source image, hereby ensuring consistency across generated images.
Conditional score guidance~\cite{lee2023conditional} derives a score function conditioned on both the source image and source prompt to guide target image generation.
Prompt interpolation-based correction~\cite{lee2025diffusion} refines the noise prediction for the target latent by progressively interpolating between the source and target prompts in a time-dependent manner.
Unlike these approaches, InstructPix2pix~\cite{brooks2023instructpix2pix} fine-tunes the pretrained Stable Diffusion using source images and generated pairs of text instructions and target images.



\section{Diffusion-Based Image-to-Image Translation}
\label{sec:diffusion_i2i}
This section describes a simple DDIM-based method tailored for the text-driven image-to-image translation task.

\subsection{Inversion process of source images}
\label{sec:background}

Diffusion models~\cite{sohl2015deep, ho2020denoising, song2020denoising} generate images through inversion and reverse processes, which correspond to the forward and backward processes, respectively.
In the inversion process, the original data $\bx_0$ is progressively perturbed with Gaussian noise, resulting in the sequence of intermediate latent variables $\bx_1, \bx_2, \dots, \bx_T$.
For text-driven image-to-image translation, existing algorithms~\cite{parmar2023zero, couairon2022diffedit} often employ the deterministic process of DDIM~\cite{song2020denoising} using pretrained text-to-image diffusion models.
The deterministic DDIM inversion process is defined as follows:
\begin{align}
	\xsrc_{t+1} =~& f^\text{inv}_t (\xsrc_t) \nonumber \\
	=~  & \sqrt{\alpha_{t+1}} \left( \frac{\xsrct - \sqrt{1-\alpha_t} \epsilon_{\theta}(\xsrct, t, \ysrc)}{\sqrt{\alpha_t}}\right) \nonumber \\
    +~& \sqrt{1-\alpha_{t+1}} \epsilon_{\theta}(\xsrct, t, \ysrc),
	\label{eq:ddim_forward}
\end{align}
where $f^\text{inv}_t(\cdot)$ denotes the inversion process at time step $t$, $\epsilon_{\theta}(\cdot, \cdot, \cdot)$ is the noise prediction network, $\xsrc_t$ is the noisy source image at time step $t$, and $\ysrc$ is the CLIP~\cite{radford2021learning} embedding of the source prompt $p^{\mathrm{src}}$.
Note that $\xsrc_T$ is obtained by recursively applying Eq.~\eqref{eq:ddim_forward}  starting from the source image $\xsrc_0$, and it is subsequently used in the reverse process to generate the target image.

\subsection{Reverse process of target images}

The target image $\xtgt_0$ is generated from $\xtgt_T$, which is set equal to $\xsrc_T$, using the DDIM reverse process given by
\begin{align}
	\xtgt_{t-1} =~&f^\text{rev}_t (\xtgt_t) \nonumber \\
	=~&\sqrt{\alpha_{t-1}} \left( \frac{\xtgtt - \sqrt{1-\alpha_t} \epsilon_{\theta}(\xtgtt, t, \ytgt)}{\sqrt{\alpha_t}}\right) \nonumber \\
    +~& \sqrt{1-\alpha_{t-1}} \epsilon_{\theta}(\xtgtt, t, \ytgt),
	\label{eq:ddim_backward}
\end{align}
where $f^\text{rev}_t(\cdot)$ is the reverse process function at time step $t$, and $\xtgt_t$ and $\ytgt$ denote the target image at time step $t$ and the CLIP feature of the target prompt $p^{\mathrm{tgt}}$, respectively.
However, recursively applying Eq.~\eqref{eq:ddim_backward} to synthesize the target image often fails to preserve the overall structure and background of the source image.
Therefore, we modify the reverse process to generate the desired target image in a training-free manner.

\begin{figure*}[t!]
	\centering
	\includegraphics[width=0.9\linewidth]{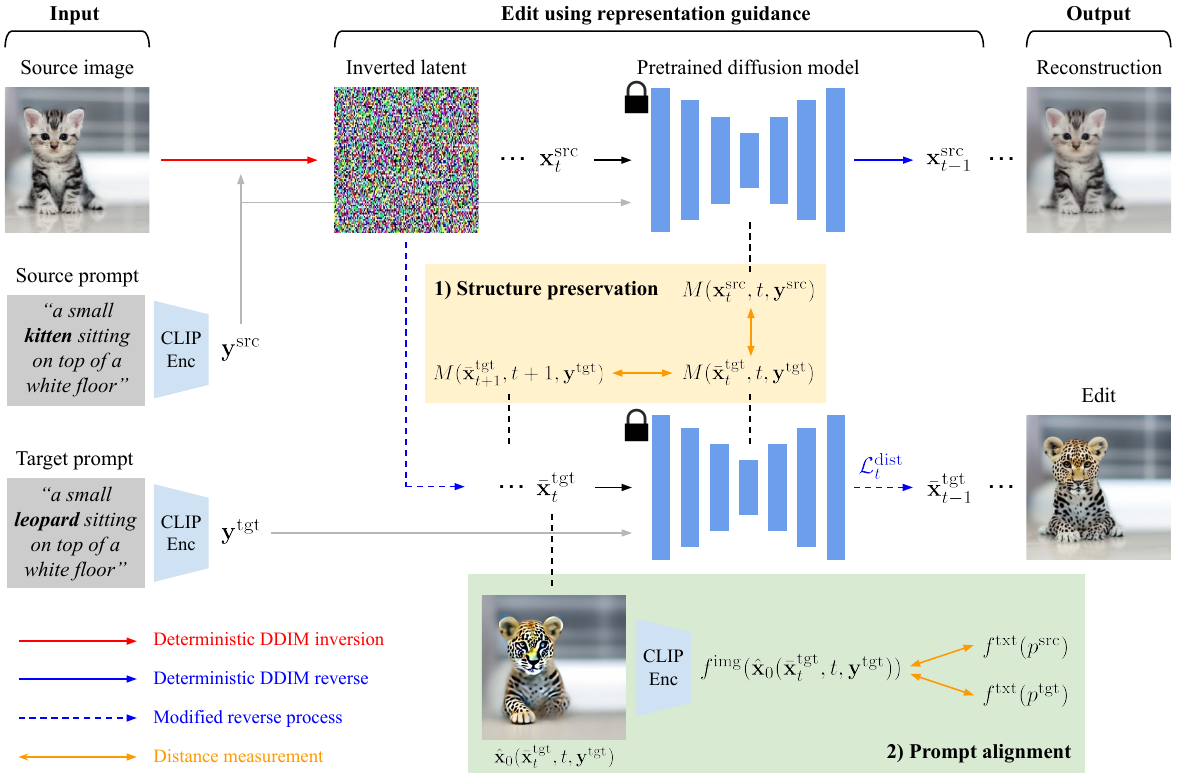}
	\caption{Overview of the proposed method about utilizing the representation guidance.}
\label{fig:method_cycle}
\end{figure*}
%



\section{Proposed Approach}
\label{sec:method}

This section elaborates on our sampling strategy, which optimizes the deterministic reverse process of DDIM with respect to the representation guidance term.

\subsection{Overview}
\label{subsec:target_image_generation}

We propose the optimized objective $\mathcal{L}^{\text{dist}}_t$, referred to as representation guidance, to address the limitations of the recursive application of Eq.~\eqref{eq:ddim_backward}, which often fails to preserve the structure of the source image.
The key idea is to regulate the reverse process, ensuring that the target image aligns semantically with the target prompt while maintaining the overall structure of the source image through the guidance term.
The modified reverse process is given by
\begin{align}
	\bxtgt_{t-1} =~&\bar{f}^\text{rev}_{t} (\bxtgt_{t}) \nonumber \\ 
	=~&\sqrt{\frac{\alpha_{t-1}}{\alpha_{t}}} \bxtgtt - \sqrt{1-\alpha_{t}} \gamma_{t} \epsilon_{\theta}(\bxtgtt, t, \ytgt)  \nonumber \\
	-~& \nabla_{\bxtgtt} \mathcal{L}^{\text{dist}}_t,
	\label{eq:bwg_sampling}
\end{align}
where $\bar{f}^\text{rev}_{t}(\cdot)$ is defined by our modified reverse process at time step $t$, and $\gamma_{t}$ is equal to $\sqrt{\frac{\alpha_{t-1}}{\alpha_{t}}}-\sqrt{\frac{1-\alpha_{t-1}}{1-\alpha_{t}}}$.  
We will discuss how $\mathcal{L}^{\text{dist}}_t$ is calculated in Section~\ref{sec:method_distance}.
Algorithm~\ref{alg:algorithm} summarizes the detailed procedures of the proposed method, referred to as Optimized Inference with Guidance (OIG).

\begin{algorithm}[t!]
	\caption{Optimized Inference with Guidance (OIG)}
	\label{alg:algorithm}
	\begin{algorithmic}
		
		\State \textbf{Inputs:} A source image $\xsrc_0$, a source prompt $\ysrc$, a target prompt $\ytgt$
        
        \For{$t \gets 0, \cdots, T-1$}
        
        \State Compute $\xsrc_{t+1}$ using Eq.~\eqref{eq:ddim_forward} 
        
        \EndFor

        \State $\bxtgt_T \gets \xsrc_T$
        
		\For{$t \gets T, \cdots , 1$}
		
        \State Compute $\hat{\bx}_0 (\bxtgt_t, t, \ytgt)$ using Eq.~\eqref{eq:tweedie}.

		\State Compute $\mathcal{L}^{\text{dist}}_t$ using Eq.~\eqref{eq:distance_triplet}.
		
		\State Compute $\gamma_{t} \gets \sqrt{\frac{\alpha_{t-1}}{\alpha_{t}}}-\sqrt{\frac{1-\alpha_{t-1}}{1-\alpha_{t}}}$.

        \State Compute $\bxtgt_{t-1}$ using Eq.~\eqref{eq:bwg_sampling}.
        
		\EndFor
		
		\State $\xtgt_0 \gets \bxtgt_0$ 
		
		\State {\bf Output:} A target image $\xtgt_0$
	\end{algorithmic}
	\label{alg1}
\end{algorithm}

\begin{table*}[h!]
	\centering
	\caption{
		Quantitative comparisons of our method with DiffEdit~\cite{couairon2022diffedit}, Plug-and-Play~\cite{tumanyan2023plug} Pix2Pix-Zero~\cite{parmar2023zero}, Null-text inversion~\cite{mokady2023null}, and MasaCtrl~\cite{cao2023masactrl} using the pretrained Stable Diffusion~\cite{rombach2022high} and images sampled from the LAION-5B dataset~\cite{schuhmann2022laion}.
		InstructPix2pix~\cite{brooks2023instructpix2pix} is included only for reference because it requires extra fine-tuning.
		For the the drawing $\rightarrow$ oil painting task, we do not report the BD score since the background is not clearly defined.
		Black and red bold-faced numbers represent the best and second-best performance in each column. 
	}
	\vspace{-2mm}
	\setlength\tabcolsep{1.8pt} 
	\scalebox{0.75}{
			\hspace{-2mm}
			\begin{tabular}{l ccc ccc ccc ccc ccc cc}
				\toprule 
				\multirow{2}{*}{Method} 
				& \multicolumn{3}{c}{cat $\rightarrow$ dog}
				& \multicolumn{3}{c}{dog $\rightarrow$ cat}
				& \multicolumn{3}{c}{dog $\rightarrow$ crochet dog}
				& \multicolumn{3}{c}{horse $\rightarrow$ zebra}  
				& \multicolumn{3}{c}{zebra $\rightarrow$ horse}  
				& \multicolumn{2}{c}{drawing $\rightarrow$ oil painting} \\
				
				\cmidrule(lr){2-4} \cmidrule(lr){5-7} \cmidrule(lr){8-10} \cmidrule(lr){11-13} \cmidrule(lr){14-16} \cmidrule(lr){17-18} 
				& CS ($\uparrow$)
				& SD ($\downarrow$)
				& BD ($\downarrow$)
				
				& CS ($\uparrow$)
				& SD ($\downarrow$)
				& BD ($\downarrow$)
				
				& CS ($\uparrow$)
				& SD ($\downarrow$)
				& BD ($\downarrow$)
				
				& CS ($\uparrow$)
				& SD ($\downarrow$)
				& BD ($\downarrow$)

				& CS ($\uparrow$)
				& SD ($\downarrow$)
				& BD ($\downarrow$)

				& \hspace{4.0mm}CS ($\uparrow$)
				& SD ($\downarrow$)

				\\ 
				\cmidrule(lr){1-18}
				
				{DiffEdit}
				&0.297  &0.025	&0.191
				&0.291	&0.033	&0.118
				&\bred{0.304}  &0.031  &0.107
				&\bred{0.322}	&0.030	&0.090
				&0.287	&0.033	&0.167
				&\hspace{4.0mm}0.278	&0.047	 	 \\

				{Plug-and-Play}
				&0.273  &\bred{0.023}  &0.224
				&0.278  &\black{0.019}  &\bred{0.106}
				&0.294  &\bred{0.020}  &0.122
				&0.310  &\black{0.022}  &\black{0.067}
				&0.275	&\bred{0.025}	&\bred{0.122}
				&\hspace{4.0mm}0.259  &0.035   \\

				{Pix2Pix-Zero}
				&\black{0.300}	&0.030	&\bred{0.189}
				&\bred{0.295}	&0.034	&0.144
				&0.303  &0.029	&0.123
				&0.316	&0.047	&0.160
				&\bred{0.289} &0.031	&0.140
				&\hspace{3mm} {0.288}  &\bred{0.023}  	 \\

				{Null-text inv.}
				&{0.296}	&0.029	&{0.200}
				&{0.294}	&0.026	&0.118
				&\bred{0.304}  &0.021	&\bred{0.093}
				&0.311	&\bred{0.025}	&0.086
				&\black{0.299}	&0.032	&0.144
				&\hspace{3mm} {0.291}  &{0.036} \\

				{MasaCtrl}
				& \bred{0.299} & 0.039 & 0.395
				& 0.293 & 0.045 & 0.225
				& 0.291 & 0.039 & 0.169
				& 0.270 & 0.033 & 0.199
				&0.276	&0.043	&0.242
				& \hspace{4.0mm}\black{0.293} & 0.073  \\

				{OIG (Ours)}
				&{0.298}	&\black{0.020}	&\black{0.153}
				&\black{0.297}	&\black{0.019}	&\black{0.084}
				&\black{0.315}	&\black{0.015}	&\black{0.071}
				&\black{0.324}	&{0.026}	&\black{0.067}
				&\bred{0.289}	&\black{0.021}	&\black{0.112}
				&\hspace{4.0mm}\black{0.293}	&\black{0.013}  \\
				
				\hdashline
				
				{InstructPix2pix}
				&0.279	&0.010	&0.129
				&0.276	&0.031	&0.107
				&0.310	&0.036	&0.154
				&0.304	&0.039	&0.139
				&0.252	&0.024	&0.311
				&\hspace{4.0mm}0.303  &0.026  	 \\
				
				\bottomrule 
			\end{tabular}
	}
	\vspace{-2mm}
	\label{tab:cmp_baselines}
\end{table*}

\subsection{Na\"ive representation guidance}
\label{sec:method_naive_distance}
During the modified reverse process, we simply derive a na\"ive distance objective, which is defined as
\begin{align}
	\mathcal{L}^{\text{naive-dist}}_t  := - & \text{Sim} \Big( f^\text{img}(\hat{\bx}_0(\bxtgt_t, t, \ytgt)), f^\text{txt}( p^{\mathrm{tgt}} ) \Big)  \nonumber \\
	+ \beta_f  \| M&(\bxtgt_t, t, \ytgt) - M(\xsrc_t, t, \ysrc)\|_F,
	\label{eq:distance_wo_triplet}
\end{align}
where $f^\text{img}(\cdot)$ and $f^\text{txt}(\cdot)$ are CLIP image and text encoders, respectively, Sim$(\cdot, \cdot)$ represents the cosine similarity between two vectors, $\| \cdot \|_F$ denotes the Frobenius norm, {$M(\cdot, \cdot, \cdot)$ extracts an intermediate feature map} from the noise prediction network $\epsilon_\theta(\cdot, \cdot, \cdot)$, and $\beta_f$ is a hyperparameter. 
Also, $\hat{\bx}_0(\cdot, \cdot, \cdot)$ is the estimated image sample based on the following Tweedie's formula~\cite{stein1981estimation}: 
\begin{equation}
	\hat{\bx}_0 (\bx_t, t, \y) = \frac{\bx_t - \sqrt{1-\alpha_t} \epsilon_{\theta}(\bx_t, t, \y)}{\sqrt{\alpha_t}}.
	\label{eq:tweedie}
\end{equation}

The second term of the right-hand side in Eq.~\eqref{eq:distance_wo_triplet} is inspired by the ability of intermediate feature maps extracted from the noise prediction network to capture the semantic information of the source image, as demonstrated in\cite{tumanyan2023plug}. 
This term promotes the preservation of semantic features from the source image in the generated target image.

\subsection{Representation guidance}
\label{sec:method_distance}

Building on the metric learning framework~\cite{hoffer2015deep}, we refine the na\"ive objective in Eq.~\eqref{eq:distance_wo_triplet} by introducing a stricter constraint. 
Specifically, we guide the generation of the target image to align more closely with the target prompt $\ptgt$ than with the source prompt $\psrc$. 
Furthermore, we encourage the generated target latent $\bar{\mathbf{x}}_t^{\text{tgt}}$ to remain semantically closer to the source latent $\mathbf{x}_t^{\text{src}}$ than $\bar{\mathbf{x}}_{t+1}^{\text{tgt}}$, thereby effectively preserving critical information from the source image.

Since the {target latent $\bar{\mathbf{x}}_{t+1}^{\text{tgt}}$ at the previous time step} inherently contains less semantic information about the source image than the source latent $\mathbf{x}_t^{\text{src}}$, we propose a stricter constraint than simply aligning $\bar{\mathbf{x}}_t^{\text{tgt}}$ with $\mathbf{x}_t^{\text{src}}$. 
Specifically, we encourage $\bar{\mathbf{x}}_t^{\text{tgt}}$ to diverge from the target latent $\bar{\mathbf{x}}_{t+1}^{\text{tgt}}$ at the previous timestep.
This design is motivated by the intuition that such a strategy better preserves the semantic content of the source image during the reverse process. 
To enforce this, we impose another constraint such that the distance between { $M(\bar{\mathbf{x}}_t^{\text{tgt}}, t, \ytgt)$ and $M(\mathbf{x}_t^{\text{src}}, t, \ysrc)$ remains smaller than the distance between $M(\bar{\mathbf{x}}_t^{\text{tgt}}, t, \ytgt)$ and $M(\bar{\mathbf{x}}_{t+1}^{\text{tgt}}, t+1, \ytgt)$. }
Empirical results show that this strategy effectively helps preserve the structure and semantic information of the source image.

In summary, we define the representation guidance via the effective distance objectives to realize the aforementioned two constraints as follows:
\begin{align}
	& {\mathcal{L}}^{\text{dist}}_t := \nonumber \\
	& - \lambda_1 \min(0, \; \text{Sim}(f^\text{img}(\hat{\bx}_0(\bxtgt_t, t, \ytgt)), f^\text{txt}(\ptgt)) \nonumber \\
	& \quad  - \text{Sim}(f^\text{img}(\hat{\bx}_0(\bxtgt_t, t, \ytgt)), f^\text{txt}(\psrc))  - \beta_p) \nonumber \\
	& + \lambda_2 \max(0, \; \beta_f \|M(\bxtgt_t, t, \ytgt)-M(\xsrc_t, t, \ysrc)\|_F  \nonumber \\
     & \quad - \|M(\bxtgt_t, t, \ytgt) - M(\bxtgt_{t+1}, t+1, \ytgt)\|_F),
	\label{eq:distance_triplet} 
\end{align}
where $\lambda_1$, $\lambda_2$, and $\beta_p$ are hyperparameters.



\section{Experiments}
\label{sec:exp}
We compare the proposed algorithm, referred to as OIG, with existing methods~\cite{couairon2022diffedit, tumanyan2023plug, parmar2023zero, mokady2023null, cao2023masactrl, brooks2023instructpix2pix}, using the pretrained Stable Diffusion~\cite{rombach2022high}.
Additionally, we present an ablation study to analyze the effects of the proposed components.

\begin{figure*}[t!]
	\centering
	\includegraphics[width=\linewidth]{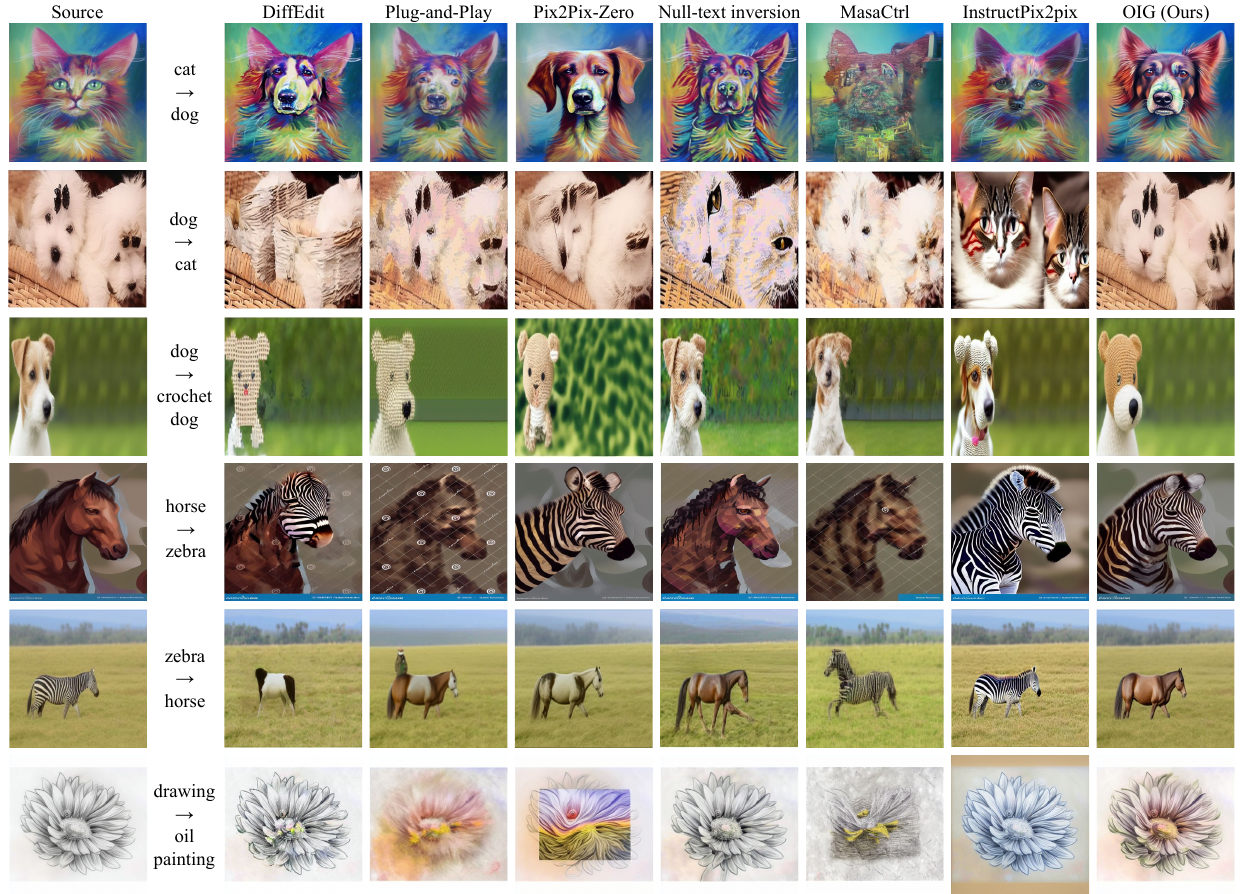}
	\caption{Qualitative comparisons between the proposed algorithm and state-of-the-art methods~\cite{couairon2022diffedit, tumanyan2023plug, parmar2023zero, mokady2023null, cao2023masactrl, brooks2023instructpix2pix} on the data sampled from the LAION-5B dataset~\cite{schuhmann2022laion} using the pretrained Stable Diffusion~\cite{rombach2022high}. 
    Note that we strictly keep the original aspect ratio of each image during all experiments, and we change the aspect ratio only for visualization. }
\label{fig:comparisons}
\end{figure*}

\subsection{Implementation details}
\label{sec:impl_detail}

We implement the proposed method using PyTorch, based on the publicly available code of Pix2Pix-Zero~\cite{parmar2023zero}.
To speed up the translation of given source images, we reduce the number of denoising timesteps to 50 for all compared algorithms including the proposed method.
Additionally, we replace the original captions with those generated by Bootstrapping Language-Image Pre-training (BLIP)~\cite{li2022blip}, as the original captions often include languages other than English.
The target prompts are constructed from the source prompts to reflect the tasks at hand.
For instance, in the cat $\rightarrow$ dog task, we replace the token most closely related to ``cat'' in the source prompt with the ``dog'' token based on the CLIP text encoder.
Note that we obtained results from the official codes of DiffEdit~\cite{couairon2022diffedit}, Plug-and-Play~\cite{tumanyan2023plug}, Pix2Pix-Zero~\cite{parmar2023zero}, Null-text inversion~\cite{mokady2023null} and MasaCtrl~\cite{cao2023masactrl} using the same source and target prompts with the classifier-free guidance~\cite{ho2021classifier} and the pretrained Stable Diffusion v1-4 checkpoint for fair comparisons.
For the InstructPix2pix~\cite{brooks2023instructpix2pix} experiments, we use the official implementation and employ text instructions corresponding to the image-to-image translation tasks, as specified in the InstructPix2pix protocol, instead of using the source and target prompts.

\subsection{Experimental settings}
\label{sec:exp_settings}

\paragraph{Dataset and tasks} 
We select about 250 images for each task from the LAION-5B dataset~\cite{schuhmann2022laion}, based on the highest CLIP similarity with the text description corresponding to the source task~\cite{beaumont-2022-clip-retrieval} to compare our method with state-of-the-art algorithms.
For evaluation, we follow the standard experimental protocols of existing methods~\cite{mokady2023null, parmar2023zero} for image-to-image translation tasks.
Specifically, we focus on object-centric tasks such as cat $\rightarrow$ dog, dog $\rightarrow$ cat, horse $\rightarrow$ zebra, and zebra $\rightarrow$ horse.
Additionally, we include the dog $\rightarrow$ crochet dog task, where the goal is to transform the dog into one that resembles a yarn figure.
We also test the proposed approach by transforming hand-drawn sketches into oil paintings in the drawing $\rightarrow$ oil painting task.

\vspace{-3mm}
\paragraph{Evaluation metrics}
We evaluate 1) how well the synthesized target image aligns with the target prompt, and 2) how effectively the structure of the source image is preserved after translation.
First, we measure the similarity between the target prompt and the generated target image using CLIP~\cite{radford2021learning}, which we call CLIP Similarity (CS). 
To assess the overall structural difference between the source and target images, we use the self-similarity map of ViT~\cite{tumanyan2022splicing} extracted from both images.
We then compute the squared Euclidean distance between their feature maps, which we call Structure Distance (SD).
Additionally, to evaluate how well each algorithm preserves the background, we identify the background components of both the source and target images by removing the object parts using the pretrained segmentation model, Detic~\cite{zhou2022detecting}.
We measure the squared Euclidean distance between the background regions of the two images, which we denote as Background Distance (BD).

\begin{table*}[t!]
	\centering
	\caption{
		Ablation study results to analyze the effect of the proposed component using the pretrained Stable Diffusion~\cite{rombach2022high} and real images sampled from the LAION-5B dataset~\cite{schuhmann2022laion} for various tasks. 
		DDIM employs the reverse process defined in Eq.~\ref{eq:ddim_backward} while Na\"ive Distance replaces the triplet-based distance objective of the representation guidance in Eq.~\ref{eq:distance_triplet} with the naive distance term in Eq.~\ref{eq:distance_wo_triplet} and Distance utilizes the representation guidance based on the triplet loss. 
	}
	\vspace{-2mm}
	\setlength\tabcolsep{1.8pt} 
	\scalebox{0.8}{
			\hspace{-2mm}
			\begin{tabular}{l ccc ccc ccc ccc ccc cc}
				\toprule 
				\multirow{2}{*}{Method} 
				& \multicolumn{3}{c}{cat $\rightarrow$ dog}
				& \multicolumn{3}{c}{dog $\rightarrow$ cat}
				& \multicolumn{3}{c}{dog $\rightarrow$ crochet dog}
				& \multicolumn{3}{c}{horse $\rightarrow$ zebra}  
				& \multicolumn{3}{c}{zebra $\rightarrow$ horse}  
				& \multicolumn{2}{c}{drawing $\rightarrow$ oil painting}\\
				
				\cmidrule(lr){2-4} \cmidrule(lr){5-7} \cmidrule(lr){8-10} \cmidrule(lr){11-13} \cmidrule(lr){14-16} \cmidrule(lr){17-18} 
				
				& CS ($\uparrow$)
				& SD ($\downarrow$)
				& BD ($\downarrow$)

				& CS ($\uparrow$)
				& SD ($\downarrow$)
				& BD ($\downarrow$)
				
				& CS ($\uparrow$)
				& SD ($\downarrow$)
				& BD ($\downarrow$)

				& CS ($\uparrow$)
				& SD ($\downarrow$)
				& BD ($\downarrow$)

				& CS ($\uparrow$)
				& SD ($\downarrow$)
				& BD ($\downarrow$)
				
				& \hspace{4.0mm}CS ($\uparrow$)
				& SD ($\downarrow$)

				\\ 
				\cmidrule(lr){1-18}
				
				{DDIM}
				&0.289	&0.072	&0.347
				&0.289	&0.063	        &0.224
				&0.299	&0.069	        &0.242		
				&0.310	&0.081	&0.255	
				&0.289	&0.079	&0.221
				&\hspace{4.0mm}0.269	&0.128	\\

				{Na\"ive Distance}
				&\black{0.298}	&\black{0.020}	&0.159
				&\black{0.297}	&0.020	&{0.087}
				&\black{0.316}	&0.017	        &0.080
				&0.322	&\black{0.023}	&0.081
				&\black{0.290}	&0.027	&0.143
				&\hspace{4.0mm}{0.292}	&0.016 \\

				{OIG (Ours)}
				&\black{0.298}	&\black{0.020}	&\black{0.153}
				&\black{0.297}	&\black{0.019}	&\black{0.084}
				&{0.315}	&\black{0.015}	&\black{0.071}
				&\black{0.324}	&0.026	         &\black{0.067}
				&0.289	&\black{0.021}	&\black{0.112}
				&\hspace{4.0mm}\black{0.293}	&\black{0.013} \\

				\bottomrule 
			\end{tabular}\
	}
	\label{tab:ablation_cycle}
\end{table*}

\begin{figure*}[t!]
	\centering
	\includegraphics[width=0.9\linewidth]{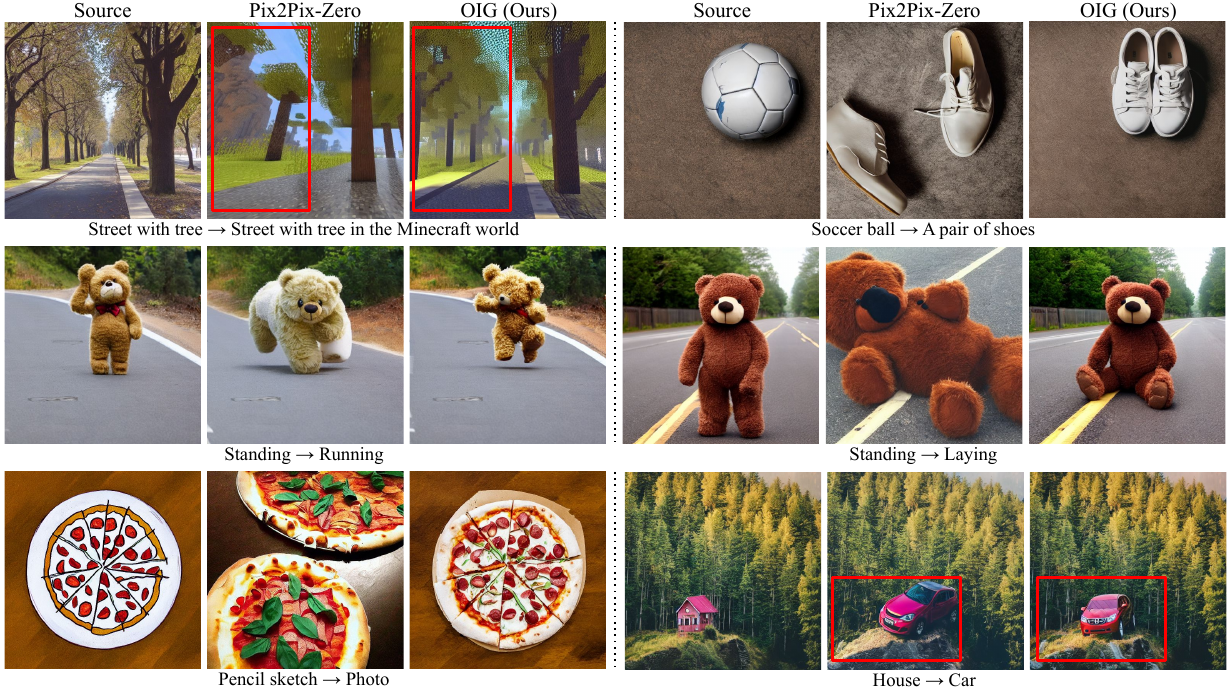}
	\caption{Qualitative comparisons between the proposed algorithm and Pix2Pix-Zero~\cite{parmar2023zero} on synthetic images given by Stable Diffusion~\cite{rombach2022high}.}
\vspace{-2mm}
\label{fig:synth_cmp_pix2pix-zero}
\end{figure*}
\begin{figure*}[t!]
	\centering
	\includegraphics[width=\linewidth]{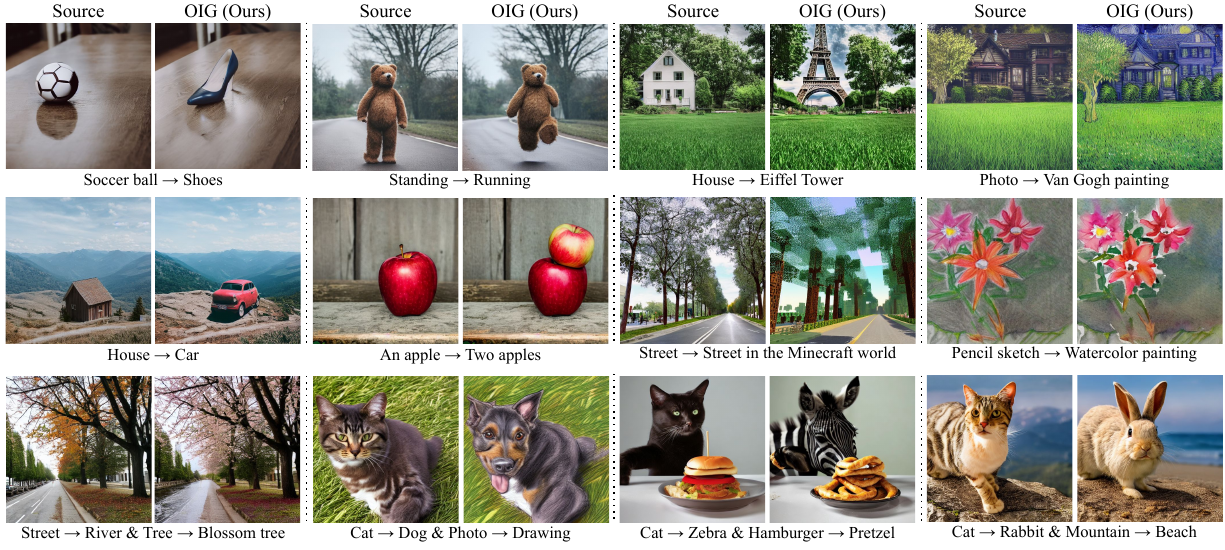}
	\vspace{-5mm}
	\caption{Editing examples of the proposed method on synthetic images given by the pretrained Stable Diffusion~\cite{rombach2022high}. Given the source and target prompts, our method generates a target image while successfully preserving the overall structure of the source image and maintaining the background excluding the parts to be manipulated.} 
\label{fig:synth_qualitative}
\end{figure*}

\subsection{Quantitative results}
\label{sec:exp_ret_quan}

We present  quantitative results in Table~\ref{tab:cmp_baselines} to compare the proposed method with state-of-the-art training-free apporaches~\cite{couairon2022diffedit, tumanyan2023plug, parmar2023zero, mokady2023null, cao2023masactrl} on various tasks using the pretrained Stable Diffusion~\cite{rombach2022high} and 250 real images sampled from the LAION-5B dataset for each task.
As shown in the table, our method outperforms the compared algorithms on most metrics, demonstrating its effectiveness for text-driven image-to-image translation tasks.
Additionally, we include the results of InstructPix2pix~\cite{brooks2023instructpix2pix}, which requires additional fine-tuning of GPT-3~\cite{brown2020language} and Stable Diffusion, unlike the proposed training-free method.
However, since the primary objective of InstructPix2pix--mapping text instructions to image-to-image translations--differs from our goal, its results are reported only for reference.

\begin{figure*}[t!]
	\centering
	\includegraphics[width=0.9\linewidth]{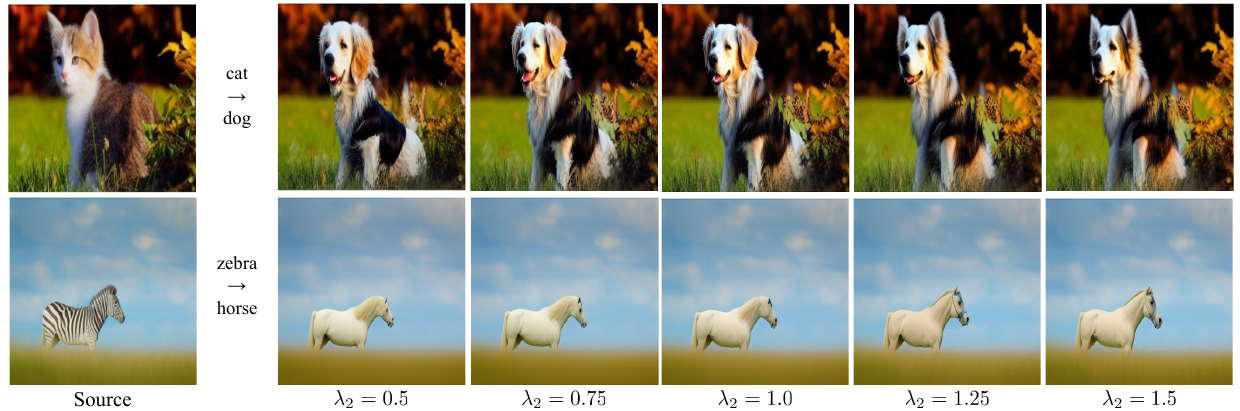}
    \caption{Qualitative reseults of the sensitivity analysis about the hyperparameter $\lambda_2$ using the pretrained Stable Diffusion~\cite{rombach2022high} and real images sampled from the LAION-5B dataset~\cite{schuhmann2022laion}.}
\vspace{-2mm}
\label{fig:ablation_hyp}
\end{figure*}

\subsection{Qualitative results}
\label{sec:exp_ret_qual}
We visualize translated results in Figure~\ref{fig:comparisons} given by the state-of-the-art methods~\cite{couairon2022diffedit, tumanyan2023plug, parmar2023zero, mokady2023null, cao2023masactrl, brooks2023instructpix2pix} and the proposed approach, all using the pretrained Stable Diffusion and images sampled from the LAION-5B dataset.
As presented in the figure, our method demonstrates superior text-driven image editing performance compared to other methods. 
Unlike our approach, competing algorithms often struggle to maintain the structural integrity of the source image in the translated outputs.

To demonstrate the superior performance of the proposed method, we qualitatively compare it with Pix2Pix-Zero using synthesized images generated by Stable Diffusion.
As shown in Figure~\ref{fig:synth_cmp_pix2pix-zero}, our method significantly outperforms prior approaches, while Pix2Pix-Zero often fails to preserve structure and introduces noticeable artifacts.
We also show additional editing examples of the proposed method on the synthesized images given by the pretrained Stable Diffusion in Figure~\ref{fig:synth_qualitative}, which verifies the effectiveness of our method. 
Notably, our method demonstrates outstanding performance even on multi-subject editing tasks.
Additional qualitative comparisons and examples are provided in Appendix E.

Furthermore, we introduce coherence guidance, which is an additional component designed to improve the fine details of the target image.
Motivated by CycleGAN~\cite{zhu2017unpaired}, we leverage the concept of cycle-consistency between the source and target domains using a pretrained diffusion model.
Empirical results demonstrate that the proposed approach effectively refines minor details in the target image generated through representation guidance, improving overall quality.
Comprehensive explanations and qualitative results of the coherence guidance are provided in Appendix B.
Note that the results shown in the main paper were obtained without incorporating the coherence guidance.

\subsection{Ablation study}
\label{sec:exp_ret_abal}
\begin{figure*}[t!]
	\centering
	\includegraphics[width=0.93\linewidth]{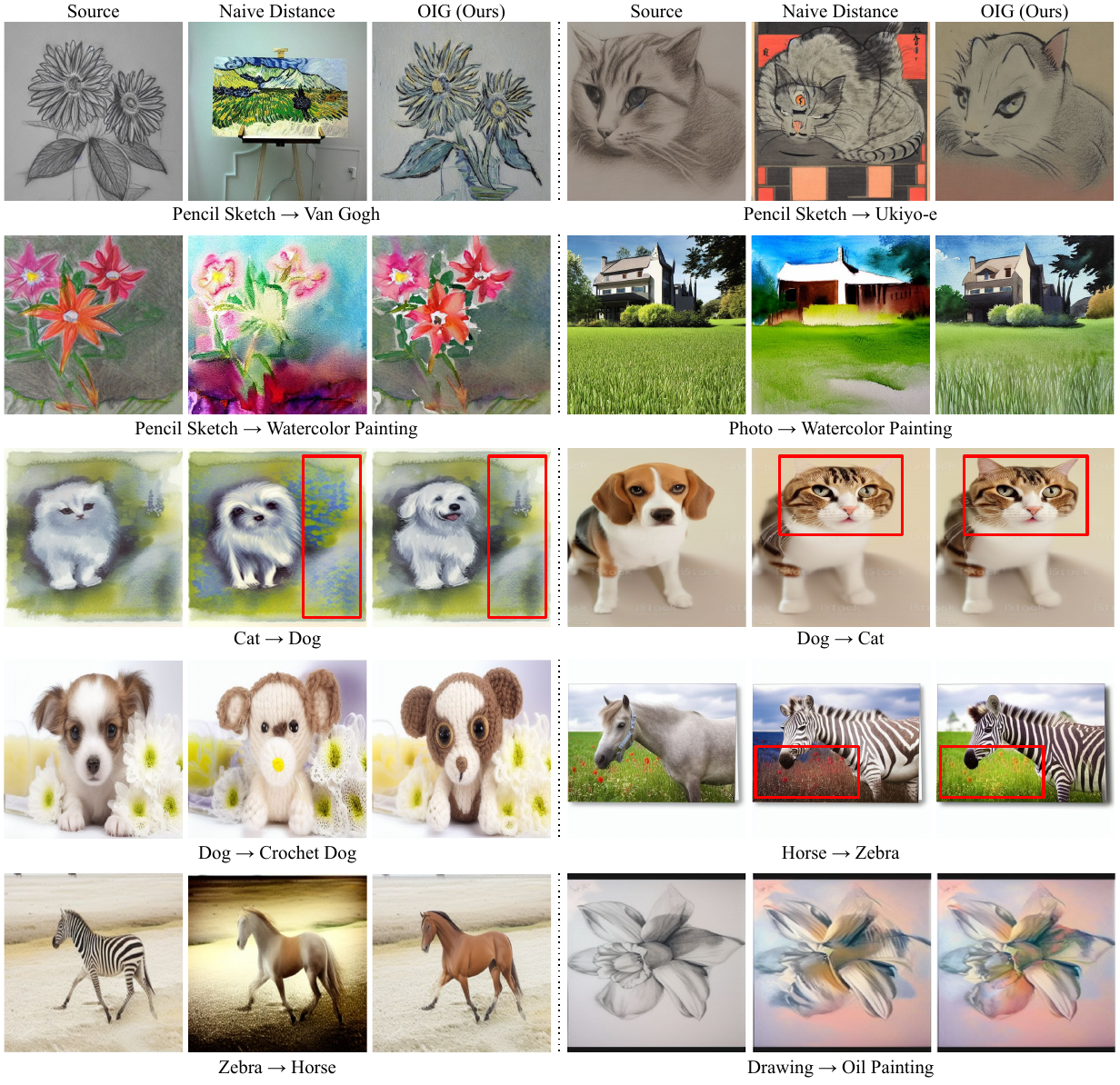}
	\caption{Qualitative results of our components on synthetic images (1st - 2nd row) and the data from LAION-5B dataset~\cite{schuhmann2022laion} (3rd - 5th row) using the pretrained Stable Diffusion~\cite{rombach2022high}. 
 Representation guidance significantly modifies the details of the target image, such as removing the artifacts, improving the backgrounds, and strictly fitting the structure into the source image.}
\label{fig:ablation}
\end{figure*}

To analyze the impact of each proposed component, we conduct an ablation study on various tasks using synthetic images generated by the pretrained Stable Diffusion and real images sampled from the LAION-5B dataset.
As demonstrated in Table~\ref{tab:ablation_cycle}, Figure~\ref{fig:ablation}, and Appendix C, the triplet-based distance term in Eq.~\eqref{eq:distance_triplet} produces target images with higher fidelity compared to the na\"ive distance objective described in Eq.~\eqref{eq:distance_wo_triplet}.

Furthermore, we qualitatively analyze the impact of the hyperparameter $\lambda_2$ on the results of the proposed method.
As illustrated in Figure~\ref{fig:ablation_hyp}, our approach exhibits robustness across a range of $\lambda_2$ values from 0.5 to 1.5.
Empirical observations reveal that smaller $\lambda_2$ values enhance alignment between the target image and the target prompt but may compromise the structural consistency with the source image.
In contrast, larger $\lambda_2$ values better preserve the structure of the source image in the generated results, albeit at the expense of slightly reduced fidelity to the target prompt.

%



\section{Conclusion}
\label{sec:conclusion}
We proposed a simple but effective method for text-driven image-to-image translation tasks based on pretrained text-to-image diffusion models.
To enhance the fidelity of the target images to the given prompt, we introduced representation guidance based on a triplet-based distance objective for the reverse process.
For the derivation of the distance objective, we encourage the target latents to align closely with the target prompts compared with the source prompts for fidelity while the target latents become closer to the source latents rather than the previous target latents for preserving the structure or background of the source image.
To demonstrate the effectiveness of our approach, we conducted extensive experiments on various image manipulation tasks and compared the results with state-of-the-art image-to-image translation methods.
Experimental results show that our framework achieves outstanding qualitative and quantitative performance in a variety of scenarios when combined with the pretrained Stable Diffusion model.

\subsection*{Acknowledgement}
This work was partly supported by the Institute of Information \& Communications Technology Planning \& Evaluation (IITP) [No.RS2022-II220959 (No.2022-0-00959), No.RS-2021-II211343, No.RS-2021-II212068] and the National Research Foundation (NRF) [No.RS-2021-NR056445 (No.2021M3A9E408078222)] funded by the Korea government (MSIT).

{\small
\bibliographystyle{ieee_fullname}
\bibliography{paper}

\begin{thebibliography}{10}\itemsep=-1pt

\bibitem{anoosheh2018combogan}
Asha Anoosheh, Eirikur Agustsson, Radu Timofte, and Luc Van~Gool.
\newblock {ComboGAN: Unrestrained Scalability for Image Domain Translation}.
\newblock In {\em CVPR}, 2018.

\bibitem{beaumont-2022-clip-retrieval}
Romain Beaumont.
\newblock {CLIP Retrieval: Easily Compute CLIP Embeddings and Build a CLIP
  Retrieval System with Them}.
\newblock \url{https://github.com/rom1504/clip-retrieval}, 2022.

\bibitem{brooks2023instructpix2pix}
Tim Brooks, Aleksander Holynski, and Alexei~A Efros.
\newblock {Instructpix2pix: Learning to Follow Image Editing Instructions}.
\newblock In {\em CVPR}, 2023.

\bibitem{brown2020language}
Tom Brown, Benjamin Mann, Nick Ryder, Melanie Subbiah, Jared~D Kaplan, Prafulla
  Dhariwal, Arvind Neelakantan, Pranav Shyam, Girish Sastry, Amanda Askell,
  et~al.
\newblock {Language Models are Few-Shot Learners}.
\newblock In {\em NeurIPS}, 2020.

\bibitem{cao2023masactrl}
Mingdeng Cao, Xintao Wang, Zhongang Qi, Ying Shan, Xiaohu Qie, and Yinqiang
  Zheng.
\newblock Masactrl: Tuning-free mutual self-attention control for consistent
  image synthesis and editing.
\newblock In {\em Proceedings of the IEEE/CVF International Conference on
  Computer Vision}, pages 22560--22570, 2023.

\bibitem{couairon2022diffedit}
Guillaume Couairon, Jakob Verbeek, Holger Schwenk, and Matthieu Cord.
\newblock {DiffEdit: Diffusion-based Semantic Image Editing with Mask
  Guidance}.
\newblock In {\em ICLR}, 2023.

\bibitem{dong2023prompt}
Wenkai Dong, Song Xue, Xiaoyue Duan, and Shumin Han.
\newblock {Prompt Tuning Inversion for Text-Driven Image Editing Using
  Diffusion Models}.
\newblock In {\em ICCV}, 2023.

\bibitem{epstein2023diffusion}
Dave Epstein, Allan Jabri, Ben Poole, Alexei Efros, and Aleksander Holynski.
\newblock Diffusion self-guidance for controllable image generation.
\newblock {\em Advances in Neural Information Processing Systems},
  36:16222--16239, 2023.

\bibitem{hertz2022prompt}
Amir Hertz, Ron Mokady, Jay Tenenbaum, Kfir Aberman, Yael Pritch, and Daniel
  Cohen-Or.
\newblock {Prompt-to-Prompt Image Editing with Cross Attention Control}.
\newblock In {\em ICLR}, 2023.

\bibitem{ho2020denoising}
Jonathan Ho, Ajay Jain, and Pieter Abbeel.
\newblock {Denoising Diffusion Probabilistic Models}.
\newblock In {\em NeurIPS}, 2020.

\bibitem{ho2021classifier}
Jonathan Ho and Tim Salimans.
\newblock {Classifier-Free Diffusion Guidance}.
\newblock In {\em NeurIPS 2021 Workshop on Deep Generative Models and
  Downstream Applications}, 2021.

\bibitem{hoffer2015deep}
Elad Hoffer and Nir Ailon.
\newblock {Deep Metric Learning using Triplet Network}.
\newblock In {\em SIMBAD}, 2015.

\bibitem{karras2017progressive}
Tero Karras, Timo Aila, Samuli Laine, and Jaakko Lehtinen.
\newblock {Progressive Growing of GANs for Improved Quality, Stability, and
  Variation}.
\newblock In {\em ICLR}, 2018.

\bibitem{kawar2023imagic}
Bahjat Kawar, Shiran Zada, Oran Lang, Omer Tov, Huiwen Chang, Tali Dekel, Inbar
  Mosseri, and Michal Irani.
\newblock {Imagic: Text-based Real Image Editing with Diffusion Models}.
\newblock In {\em CVPR}, 2023.

\bibitem{kim2022diffusionclip}
Gwanghyun Kim, Taesung Kwon, and Jong~Chul Ye.
\newblock {DiffusionCLIP: Text-guided Diffusion Models for Robust Image
  Manipulation}.
\newblock In {\em CVPR}, 2022.

\bibitem{kingma2014autoencoding}
Diederik~P. Kingma and Max Welling.
\newblock {Auto-Encoding Variational Bayes}.
\newblock In {\em ICLR}, 2014.

\bibitem{lee2023conditional}
Hyunsoo Lee, Minsoo Kang, and Bohyung Han.
\newblock {Conditional Score Guidance for Text-Driven Image-to-Image
  Translation}.
\newblock In {\em NeurIPS}, 2023.

\bibitem{lee2025diffusion}
Junsung Lee, Minsoo Kang, and Bohyung Han.
\newblock Diffusion-based image-to-image translation by noise correction via
  prompt interpolation.
\newblock In {\em ECCV}, 2025.

\bibitem{li2022blip}
Junnan Li, Dongxu Li, Caiming Xiong, and Steven Hoi.
\newblock {BLIP: Bootstrapping Language-Image Pre-training for Unified
  Vision-Language Understanding and Generation}.
\newblock In {\em ICML}, 2022.

\bibitem{mokady2023null}
Ron Mokady, Amir Hertz, Kfir Aberman, Yael Pritch, and Daniel Cohen-Or.
\newblock {Null-text Inversion for Editing Real Images Using Guided Diffusion
  Models}.
\newblock In {\em CVPR}, 2023.

\bibitem{parmar2023zero}
Gaurav Parmar, Krishna~Kumar Singh, Richard Zhang, Yijun Li, Jingwan Lu, and
  Jun-Yan Zhu.
\newblock {Zero-Shot Image-to-Image Translation}.
\newblock In {\em SIGGRAPH}, 2023.

\bibitem{radford2021learning}
Alec Radford, Jong~Wook Kim, Chris Hallacy, Aditya Ramesh, Gabriel Goh,
  Sandhini Agarwal, Girish Sastry, Amanda Askell, Pamela Mishkin, Jack Clark,
  et~al.
\newblock {Learning Transferable Visual Models from Natural Language
  Supervision}.
\newblock In {\em ICML}, 2021.

\bibitem{ramesh2022hierarchical}
Aditya Ramesh, Prafulla Dhariwal, Alex Nichol, Casey Chu, and Mark Chen.
\newblock {Hierarchical Text-Conditional Image Generation with CLIP Latents}.
\newblock {\em arXiv preprint arXiv:2204.06125}, 2022.

\bibitem{rombach2022high}
Robin Rombach, Andreas Blattmann, Dominik Lorenz, Patrick Esser, and Bj{\"o}rn
  Ommer.
\newblock {High-Resolution Image Synthesis with Latent Diffusion Models}.
\newblock In {\em CVPR}, 2022.

\bibitem{saharia2022photorealistic}
Chitwan Saharia, William Chan, Saurabh Saxena, Lala Li, Jay Whang, Emily~L
  Denton, Kamyar Ghasemipour, Raphael Gontijo~Lopes, Burcu Karagol~Ayan, Tim
  Salimans, et~al.
\newblock {Photorealistic Text-to-Image Diffusion Models with Deep Language
  Understanding}.
\newblock In {\em NeurIPS}, 2022.

\bibitem{schuhmann2022laion}
Christoph Schuhmann, Romain Beaumont, Richard Vencu, Cade Gordon, Ross
  Wightman, Mehdi Cherti, Theo Coombes, Aarush Katta, Clayton Mullis, Mitchell
  Wortsman, et~al.
\newblock {LAION-5B: An Open Large-Scale Dataset for Training Next Generation
  Image-Text Models}.
\newblock In {\em NeurIPS Datasets and Benchmarks Track}, 2022.

\bibitem{sohl2015deep}
Jascha Sohl-Dickstein, Eric Weiss, Niru Maheswaranathan, and Surya Ganguli.
\newblock {Deep Unsupervised Learning using Nonequilibrium Thermodynamics}.
\newblock In {\em ICML}, 2015.

\bibitem{song2020denoising}
Jiaming Song, Chenlin Meng, and Stefano Ermon.
\newblock {Denoising Diffusion Implicit Models}.
\newblock In {\em ICLR}, 2021.

\bibitem{stein1981estimation}
Charles~M Stein.
\newblock {Estimation of the Mean of a Multivariate Normal Distribution}.
\newblock {\em The annals of Statistics}, 1981.

\bibitem{su2022dual}
Xuan Su, Jiaming Song, Chenlin Meng, and Stefano Ermon.
\newblock {Dual Diffusion Implicit Bridges for Image-to-Image Translation}.
\newblock In {\em ICLR}, 2022.

\bibitem{tumanyan2022splicing}
Narek Tumanyan, Omer Bar-Tal, Shai Bagon, and Tali Dekel.
\newblock {Splicing ViT Features for Semantic Appearance Transfer}.
\newblock In {\em CVPR}, 2022.

\bibitem{tumanyan2023plug}
Narek Tumanyan, Michal Geyer, Shai Bagon, and Tali Dekel.
\newblock {Plug-and-play Diffusion Features for Text-Driven Image-to-Image
  Translation}.
\newblock In {\em CVPR}, 2023.

\bibitem{valevski2022unitune}
Dani Valevski, Matan Kalman, Yossi Matias, and Yaniv Leviathan.
\newblock {Unitune: Text-Driven Image Editing by Fine-tuning an Image
  Generation Model on a Single Image}.
\newblock In {\em SIGGRAPH}, 2023.

\bibitem{zhou2022detecting}
Xingyi Zhou, Rohit Girdhar, Armand Joulin, Philipp Kr{\"a}henb{\"u}hl, and
  Ishan Misra.
\newblock {Detecting Twenty-thousand Classes using Image-level Supervision}.
\newblock In {\em ECCV}, 2022.

\bibitem{zhu2017unpaired}
Jun-Yan Zhu, Taesung Park, Phillip Isola, and Alexei~A Efros.
\newblock {Unpaired Image-to-Image Translation using Cycle-Consistent
  Adversarial Networks}.
\newblock In {\em CVPR}, 2017.

\end{thebibliography}
}

\clearpage



\appendix
\section*{Appendix}

In the Appendix, we analyze the runtime and computational complexity of our method and compare it with state-of-the-art methods~\cite{couairon2022diffedit, tumanyan2023plug, parmar2023zero, mokady2023null, cao2023masactrl} in Section~\ref{sec:appendix_analysis}.
Section~\ref{sec:appendix_coherence_guidance} introduces additional component, referred to as coherence guidance, which can be combined with the proposed method to enhance the quality of the target image. 
In section~\ref{sec:appendix_ablation}, we visualize additional ablation study results.
Section~\ref{sec:appendix_pretrained_model} demonstrates the qualitative results of our method combined with other pretrained diffusion models other than Stable Diffusion~\cite{rombach2022high} to highlight the generalizability of the proposed method.
Additional qualitative results are provided in Section~\ref{sec:appendix_qualitative}.
Finally, we discuss the limitations and potential social impacts of the proposed method in Section~\ref{sec:appendix_limitations} and~\ref{sec:appendix_social}, respectively.

\section{Analysis on computational complexity}
\label{sec:appendix_analysis}

In Table~\ref{tab:cmp_complexity}, we report the runtime and computational complexity of the proposed method and state-of-the-art methods~\cite{couairon2022diffedit, tumanyan2023plug, parmar2023zero, mokady2023null, cao2023masactrl} analyzed on a single NVIDIA A100 GPU.
As shown in the table, our method shows comparable computational cost to prior works.
Since the proposed method shows superior performance compared to prior works, this demonstrates that our method is a simple but effective approach.

\begin{table*}[h!]
	\centering
	\caption{
		Runtime and computational complexity analysis of DiffEdit~\cite{couairon2022diffedit}, Plug-and-Play~\cite{tumanyan2023plug} Pix2Pix-Zero~\cite{parmar2023zero}, Null-text inversion~\cite{mokady2023null}, MasaCtrl~\cite{cao2023masactrl} and the proposed method.
        Each algorithm is tested on a single NVIDIA A100 GPU.
        The proposed method achieves comparable runtime and memory consumption compared to prior works, while outperforming prior works. }
	\setlength\tabcolsep{3.0pt} 
	\scalebox{0.85}{
			\hspace{-2mm}
			\begin{tabular}{c c c c c c c}
				\toprule 
				\multirow{1}{*}{Method} 
				& \multicolumn{1}{c}{DiffEdit}
				& \multicolumn{1}{c}{Plug-and-Play}
				& \multicolumn{1}{c}{Pix2Pix-Zero}
				& \multicolumn{1}{c}{Null-text inversion}  
				& \multicolumn{1}{c}{MasaCtrl}  
				& \multicolumn{1}{c}{OIG (Ours)} \\

				\cmidrule(lr){1-7}
				
				{time/image (s)}
				&18.24  &19.78  &30.53   &105.80  & 16.08 &37.89   \\\

				{GPU Memory (GB)}
				&4.188  &4.103	&11.975   &10.110	& 10.68 &8.363	 \\

				\bottomrule 
			\end{tabular}
	}
	\label{tab:cmp_complexity}
\end{table*}

\section{Discussion on coherence guidance}
\label{sec:appendix_coherence_guidance}

\subsection{Revised target generation}
\label{subsec:appendix_target_image_generation}
\begin{figure*}[h!]
	\centering
	\includegraphics[width=0.9\linewidth]{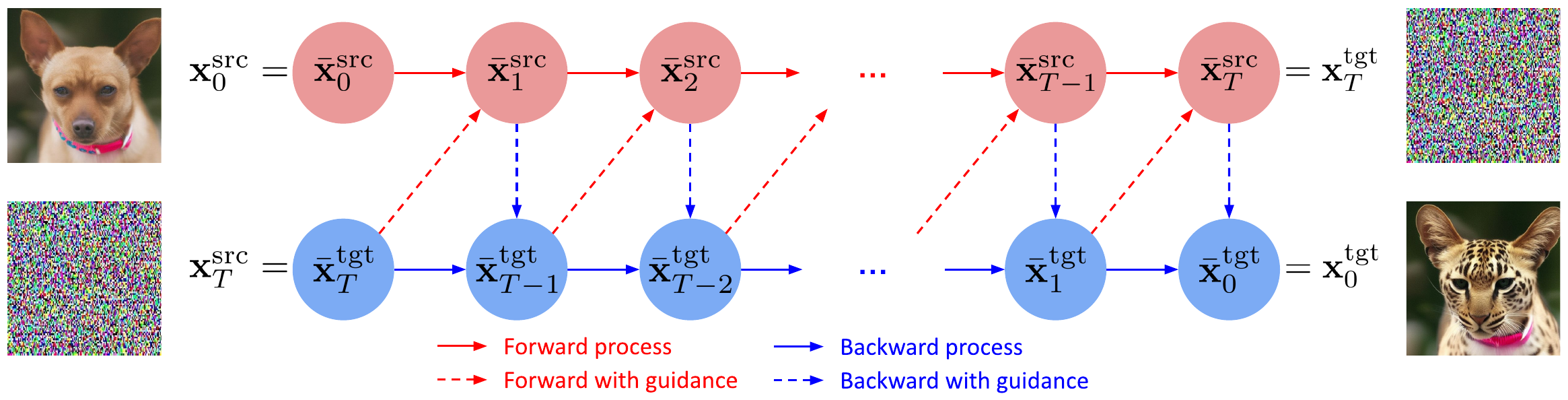}
	\caption{Overview of the revised method about the forward guidance and backward guidance.}
\label{fig:appendix_method_cycle}
\end{figure*}

Different from previous frameworks~\cite{hertz2022prompt, tumanyan2023plug, couairon2022diffedit} that revise the backward process only, the revised method alternates the estimation of the source and target latents in the order of $\{ \bxsrc_{T-t}, \bxtgtt \}_{t=T-1:0}$, where $\bxsrc_{T-t}$ and $\bxtgt_{t}$ are source and target latents obtained from our modified forward and backward processes as shown in Figure~\ref{fig:appendix_method_cycle}. 
Note that, in the case of $t=T$, $\bxsrc_{0}$ is equal to $\xsrc_0$, which is the source image, while $\bxtgt_T$ is set to $\xsrc_T$, which is given by recursively performing the deterministic DDIM inversion from the source image.
The two modified processes are denoted by \textit{forward with guidance} and \textit{backward with guidance}. 
We refer our revised method to Optimized Inference with Guidance$^{+}$ (OIG$^{+}$).
Algorithm~\ref{alg:algorithm_2} summarizes the detailed procedures of the proposed guidances.

\subsubsection{Forward with guidance}
\label{subsubsec:forward_with_guidance}
We revise the forward process in Eq.~\eqref{eq:ddim_forward} by additionally optimizing the proposed efficient version of the cycle-consistency objective $\mathcal{L}^{\text{cycle, eff}}$ as described in Section~\ref{sec:efficient_cycle_distance} with respect to the source latent $\bxsrc_{T-t}$ as follows:
\begin{align}
	\bxsrc_{T-t+1} = &\bar{f}^\text{fwd}_{T-t} (\bxsrc_{T-t}) \nonumber \\
	= &\sqrt{\frac{\alpha_{T-t+1}}{\alpha_{T-t}}} \bxsrc_{T-t} \nonumber \\ 
	- & \sqrt{1-\alpha_{T-t}} \gamma'_{T-t} \epsilon_{\theta}(\bxsrc_{T-t}, T-t, \ysrc)  \nonumber \\
	- & \nabla_{\bxsrc_{T-t}}\lambda_3 \mathcal{L}^{\text{cycle, eff}},
	\label{eq:fwg_sampling_cycle}
\end{align}
where $\bar{f}^\text{fwd}_{T-t}(\cdot)$ denotes the modified forward process at time step $T-t$, $\gamma'_{T-t}$ is equal to $\sqrt{\frac{\alpha_{T-t+1}}{\alpha_{T-t}}}- \sqrt{\frac{1-\alpha_{T-t+1}}{1-\alpha_{T-t}}}$,  and $\lambda_3$ is a hyperparameter.

\subsubsection{Backward with guidance}
In the backward with guidance process, we improve the backward process in Eq.~\eqref{eq:ddim_backward} by optimizing both the distance term $\mathcal{L}^{\text{dist}}_t$ and the cycle-consistency objective $\mathcal{L}^{\text{cycle}}$, where the modified backward process is given by
\begin{align}
	\bxtgt_{t-1} = &\bar{f}^\text{bwd}_{t} (\bxtgt_{t}) \nonumber \\ 
	= &\sqrt{\frac{\alpha_{t-1}}{\alpha_{t}}} \bxtgtt - \sqrt{1-\alpha_{t}} \gamma_{t} \epsilon_{\theta}(\bxtgtt, t, \ytgt)  \nonumber \\
    - & \nabla_{\bxtgtt}(\mathcal{L}^{\text{dist}}_t + \lambda_4 \mathcal{L}^{\text{cycle}}),
    \label{eq:bwg_sampling_cycle}
\end{align}
where $\bar{f}^\text{bwd}_{t}(\cdot)$ is defined by our modified backward process at time step $t$, $\gamma_{t}$ is equal to $\sqrt{\frac{\alpha_{t-1}}{\alpha_{t}}}-\sqrt{\frac{1-\alpha_{t-1}}{1-\alpha_{t}}}$, and $\lambda_4$ is a hyperparameter. 
$\mathcal{L}^{\text{dist}}_t$ is the distance objective defined in Eq.~\eqref{eq:distance_triplet}.
We will discuss $\mathcal{L}^{\text{cycle}}$ in Section~\ref{sec:cycle_distance}.

\begin{algorithm}[t!]
	\caption{Text-Driven Image Editing based on Forward and Backward Guidances}
	\label{alg:algorithm_2}
	\begin{algorithmic}

		\State \textbf{Inputs:} A source image $\xsrc_0$, a source prompt $\ysrc$, a target prompt $\ytgt$
		
		\For{$t \gets 0, \cdots, T-1$}
        
        \State Compute $\xsrc_{t+1}$ using Eq.~\eqref{eq:ddim_forward}
        
        \EndFor
		
		\State $\bxtgt_T \gets \xsrc_T$ and $\bxsrc_0 \gets \xsrc_0$
		
		\For{$t \gets T, \cdots , 1$}

		\State Calculate $\mathcal{L}^{\text{cycle, eff}}$ using Eq.~\eqref{eq:cycle_consistency_term_efficient} 
		
		\State Compute $\gamma'_{T-t} \gets \sqrt{\frac{\alpha_{T-t+1}}{\alpha_{T-t}}}- \sqrt{\frac{1-\alpha_{T-t+1}}{1-\alpha_{T-t}}}$
		
        \State Calculate $\bxsrc_{T-t+1}$ using Eq.~\eqref{eq:fwg_sampling_cycle}  
   
        \Comment{Forward with guidance \\}

        \State Calculate $\hat{\bx}_0 (\bxtgt_t, t, \ytgt)$ using Eq.~\eqref{eq:tweedie}
  
		\State Calculate $\mathcal{L}^{\text{dist}}_t$ using Eq.~\eqref{eq:distance_triplet}
		
		\State Compute $\mathcal{L}^{\text{cycle}}$ using Eq.~\eqref{eq:cycle_consistency_term}
		
		\State Compute $\gamma_{t} \gets \sqrt{\frac{\alpha_{t-1}}{\alpha_{t}}}-\sqrt{\frac{1-\alpha_{t-1}}{1-\alpha_{t}}}$
		
		\State Compute $\bxtgt_{t-1}$ using Eq.~\eqref{eq:bwg_sampling_cycle} 
  
        \Comment{Backward with guidance}
		
		\EndFor
		
		\State $\xtgt_0 \gets \bxtgt_0$ 
		
		\State {\bf Output:} A target image $\xtgt_0$
		
	\end{algorithmic}
\end{algorithm}

\subsection{Coherence guidance via cycle-consistency}
\label{sec:cycle_distance}

The \textit{simple DDIM translation}, recursively using the backward process defined in Eq.~\eqref{eq:ddim_backward} from the final target latent $\xtgt_T$, guarantees the cycle-consistency property as verified by~\cite{su2022dual}.
In other words, after converting the source domain image $\xsrc_0$ into $\xtgt_0$ in the target domain and then transforming $\xtgt_0$ back to the source domain image denoted by $\mathbf{\hat{x}}^{\text{src}}_0$, the equality $\xsrc_0 = \mathbf{\hat{x}}^{\text{src}}_0$ holds.

Although the simple DDIM translation guarantees the cycle-consistency, the property fails to hold in OIG because the generation process is modified by incorporating the representation guidance.
Hence, we add an objective to enforce the cycle-consistency to further enhance translation results.
As described in CycleGAN~\cite{zhu2017unpaired}, the cycle-consistency term is defined as $\| \xsrc_0 - h(g(\xsrc_0)) \|_{2,2} $ in principle, 
where $g(\cdot)$ is the image-to-image translation operation from the source domain to the target domain, and vise versa for $h(\cdot)$. 
However, with the assumption that $g(\cdot)$ and $h(\cdot)$ are invertible,
we alternatively optimize the following cycle-consistency objective, which is given by
\begin{equation}
	\mathcal{L}^{\text{cycle}} := \| \bxtgt_{0, f} - \bxtgt_{0, b}  \|_{2,2},
	\label{eq:cycle_consistency_term}
\end{equation}
where we denote $\bxtgt_{0, f}$ by $h^{-1}(\xsrc_0)$ and $\bxtgt_{0, b}$ by $g(\xsrc_0)$. 
The definition of $h^{-1}(\cdot)$ and $g(\cdot)$ are given by
\begin{align}
    \bxtgt_{0, f} &= h^{-1}(\xsrc_0)  = F^\text{bwd} (\bar{F}^\text{fwd}(\xsrc_0)) \nonumber \\ 
    \bxtgt_{0, b} &= g(\xsrc_0) = \bar{F}^\text{bwd} (F^\text{fwd}(\xsrc_0)),
    \label{eq:g_and_h_def}
\end{align}
where the auxiliary functions are defined as
\begin{align}
    F^\text{fwd}(\cdot) &=  f^\text{fwd}_{T-1} \circ f^\text{fwd}_{T-2} \cdots \circ f^\text{fwd}_{0}(\cdot), \nonumber \\
    F^\text{bwd}(\cdot) &= f^\text{bwd}_{1} \circ f^\text{bwd}_{2} \cdots  \circ f^\text{bwd}_{T} (\cdot) \nonumber \\
    \bar{F}^\text{fwd}(\cdot) &= \bar{f}^\text{fwd}_{T-1} \circ \bar{f}^\text{fwd}_{T-2} \cdots  \circ \bar{f}^\text{fwd}_{0} (\cdot) \nonumber \\
    \bar{F}^\text{bwd}(\cdot) &= \bar{f}^\text{bwd}_{1} \circ \bar{f}^\text{bwd}_{2} \cdots  \circ \bar{f}^\text{bwd}_{T} (\cdot). \nonumber
\end{align}

\paragraph{Estimation of $\bxtgt_{0, f}$} 
Using the equivalent ordinary differential equation of the simple DDIM forward process in Eq~\eqref{eq:ddim_forward}, we first approximate $\bxtgt_{T, f}$ which is equal to $\bxsrc_T$ as
\begin{align}
	\bxtgt_{T, f}&=\bxsrc_T \approx \sqrt{\frac{\alpha_T}{\alpha_{t_0}}} {\bar{\mathbf{x}}}^{\text{src}}_{t_0} \nonumber \\
    & \hspace{-1.0cm} +   \left( \sqrt{1-\alpha_T}   -\sqrt{\frac{\alpha_T (1-\alpha_{t_0})}{\alpha_{t_0}}}  \right) \epsilon_{\theta}({\bar{\mathbf{x}}}^{\text{src}}_{t_0},t_0, \ysrc),
	\label{eq:cycle_fwg_naive}
\end{align}
where $t_0$ is an intermediate time step.
Although the approximation incurs discretization errors due to the one-step estimation of ${\bar{\mathbf{x}}}^{\text{src}}_{T}$ from ${\bar{\mathbf{x}}}^{\text{src}}_{t_0}$, we empirically observe that the proposed method achieves remarkable performance as demonstrated in Section~\ref{sec:appendix_cycle_result}.
When we estimate $\bxtgt_{t-1}$ from $\bxtgt_{t}$ during the backward with guidance, $\bxtgt_{T, f}$ is derived from Eq.~\eqref{eq:cycle_fwg_naive} by plugging $T-t+1$ into $t_0$.
Finally, $\bxtgt_{0, f}$ is obtained by $F^\text{bwd}(\bxtgt_{T, f})$, where $F^\text{bwd}(\cdot)$ performs $T$ steps of the recursive backward process in Eq.~\eqref{eq:ddim_backward}.

\begin{figure*}[h!]
	\centering
	\includegraphics[width=1.0\linewidth]{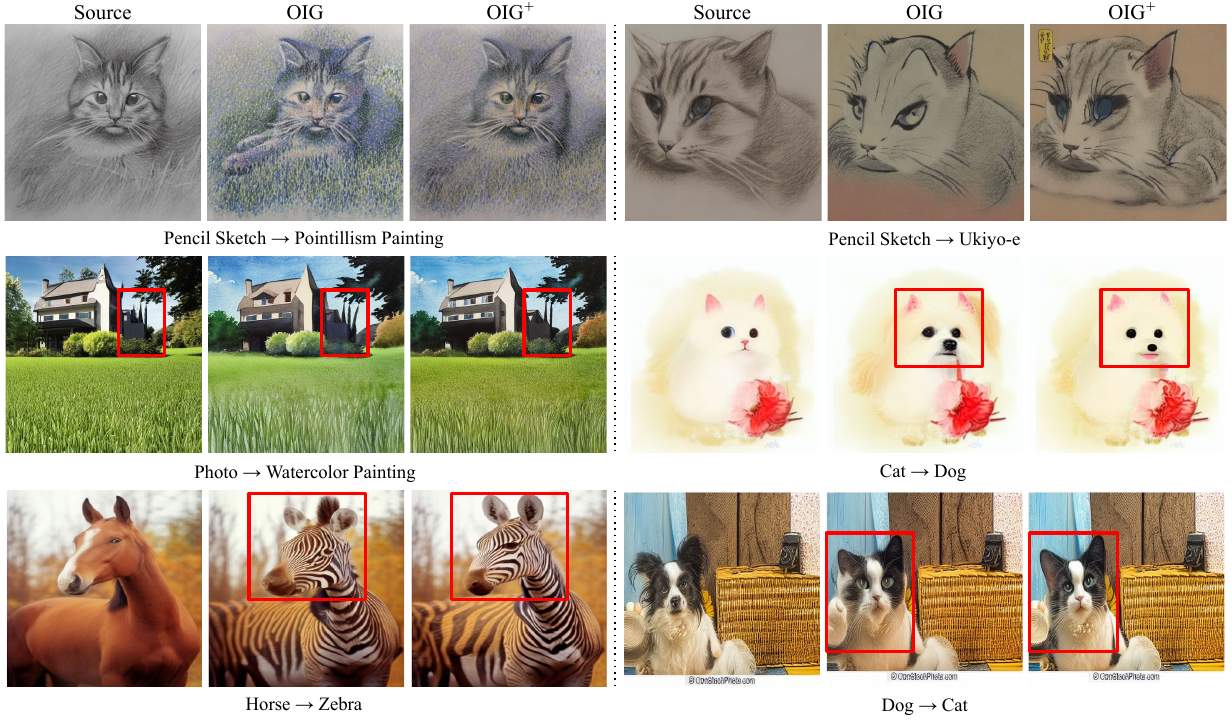}
	\caption{Qualitative results of coherence guidance on the data from LAION-5B dataset~\cite{schuhmann2022laion} and synthetic images using the pretrained Stable Diffusion~\cite{rombach2022high}. Coherence guidance effectively modifies the details of the target image when combined with OIG.}
	\label{fig:OIG*_supple}
\end{figure*}

\paragraph{Estimation of $\bxtgt_{0, b}$} 
To reduce the computational costs for the estimation of $\bxtgt_{0, b}$ from $\bxtgtt$, we approximate $\bxtgt_{0, b}$ using the Tweedie's formula~\cite{stein1981estimation} in Eq.~\eqref{eq:tweedie} as 
\begin{equation}
	\bxtgt_{0, b} \approx \hat{\bx}_0 (\bxtgt_t, t, \ytgt).
	\label{eq:cycle_bwg_tweedie}
\end{equation}
We eventually calculate $\mathcal{L}^{\text{cycle}}$ by plugging $\bxtgt_{0, f}$ and $\bxtgt_{0, b}$ into Eq.~\eqref{eq:cycle_consistency_term}.

\subsection{Efficient coherence guidance}
\label{sec:efficient_cycle_distance}

In the case of the forward with guidance, computing the gradient of $\mathcal{L}^{\text{cycle}}$ in Eq.~\eqref{eq:cycle_consistency_term} with respect to the source latent $\bxsrc_{T-t}$, denoted as $\nabla_{\bxsrc_{T-t}} \mathcal{L}^{\text{cycle}}$, is memory-intensive and time-consuming since it involves multiple times of backpropagation through the noise prediction network.
To tackle this issue, we alternatively derive the following efficient version of the cycle-consistency objective that matches the final target latents instead of the target images as
\begin{equation}
	\mathcal{L}^{\text{cycle, eff}} := \| \bxtgt_{T, f} - \bxtgt_{T, b}  \|_{2,2}.
	\label{eq:cycle_consistency_term_efficient}
\end{equation}
In the above equation, $\bxtgt_{T, f}$ is obtained based on a single forward propagation of the noise prediction network from Eq.~\eqref{eq:cycle_fwg_naive} by setting $t_0=T-t$.
We obtain $\bxtgt_{T, b}$ from $F^\text{fwd}(\bxtgt_{0, b})$, where $\bxtgt_{0, b}$ is estimated using Eq.~\eqref{eq:cycle_bwg_tweedie} and $F^\text{fwd}(\cdot)$ recursively applies the DDIM inversion process in Eq.~\eqref{eq:ddim_forward} for $T$ times.

Therefore, we can compute the gradient of $\mathcal{L}^{\text{cycle, eff}}$ with respect to $\bxsrc_{T-t}$ just by performing a single backpropagation through the noise prediction network.

\subsection{Qualitative results of OIG$^{+}$}
\label{sec:appendix_cycle_result}

We visualize the effect of coherence guidance in Figure~\ref{fig:OIG*_supple}. 
As shown in the Figure, OIG$^{+}$ enhances the fine details of the target image generated by OIG, such as reducing high-frequency noise or facilitating the alignment of small structural elements.

\section{Additional ablation study}
\label{sec:appendix_ablation}

\begin{figure*}[h!]
	\centering
	\includegraphics[width=1.0\linewidth]{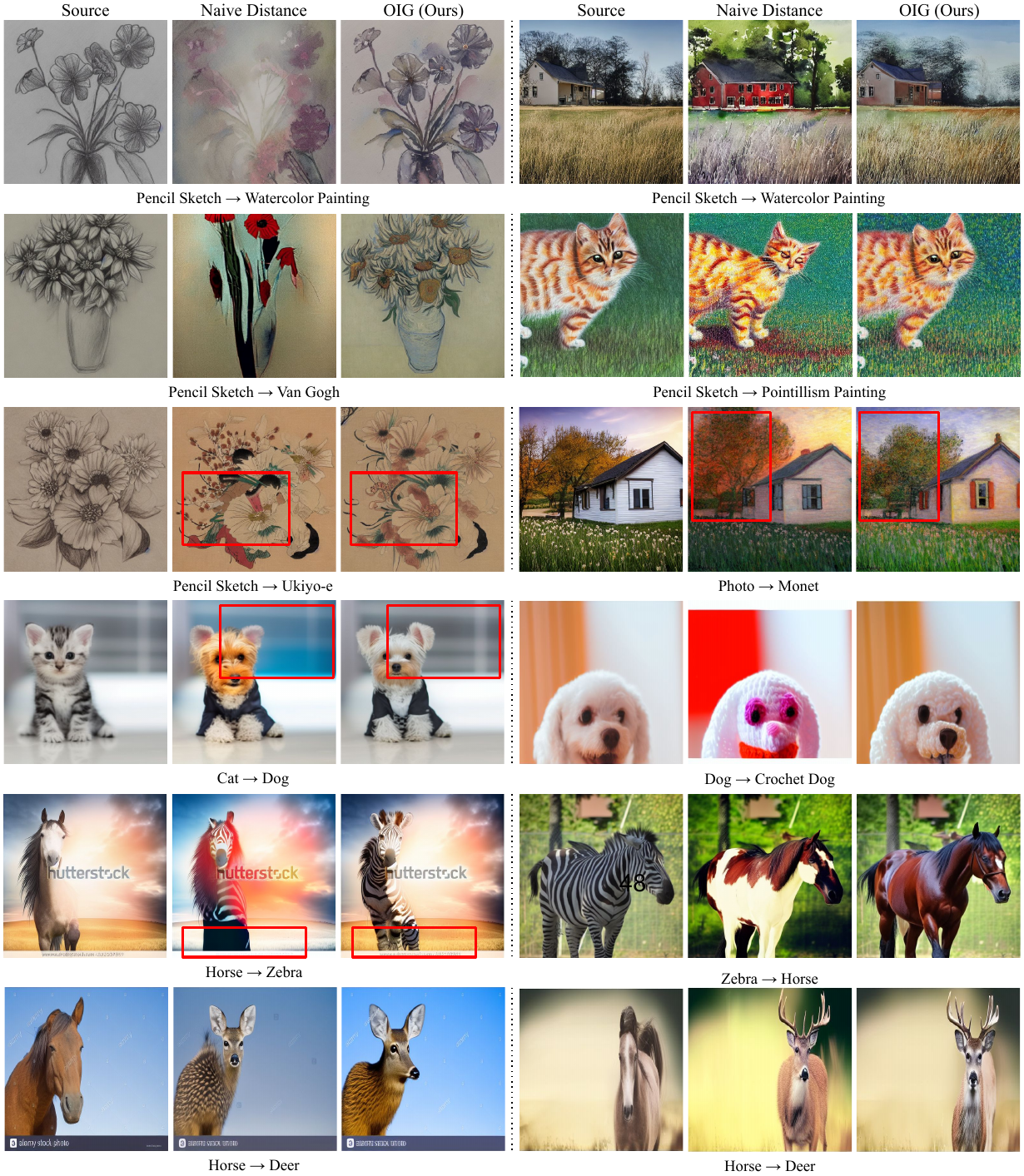}
	\caption{Qualitative results of our component on the data from LAION-5B dataset~\cite{schuhmann2022laion} and synthetic images using the pretrained Stable Diffusion~\cite{rombach2022high}. Representation guidance significantly improves the details of the target image, such as correcting the structural inconsistencies between source and target images, preserving the structure of foreground region, and enhancing the fidelity.}
	\label{fig:ablation_supple}
\end{figure*}

We report additional ablation study results of the proposed method in Figure~\ref{fig:ablation_supple}. 
We emphasize that the triplet-based distance term in Eq.~\eqref{eq:distance_triplet} enhances the fidelity of the target image and preserves the overall structure well compared to the na\"ive distance objective in Eq.~\eqref{eq:distance_wo_triplet}.

\section{Additional results using other pretrained diffusion models}
\label{sec:appendix_pretrained_model}

To demonstrate that the proposed method generalizes well to other pretrained models, we generated target images using our method with pretrained Distilled Stable Diffusion\footnote{\url{https://huggingface.co/docs/diffusers/en/using-diffusers/distilled_sd}} and Latent Diffusion Model (LDM)~\cite{rombach2022high}.
Note that Distilled Stable Diffusion is a lightweight model that has been trained by reducing the parameters of the denoising U-Net.
Also, the pipeline of LDM is similar to Stable Diffusion, however the resolution of training data for LDM differ from those of Stable Diffusion.

\begin{figure*}[h!]
	\centering
	\includegraphics[width=1.0\linewidth]{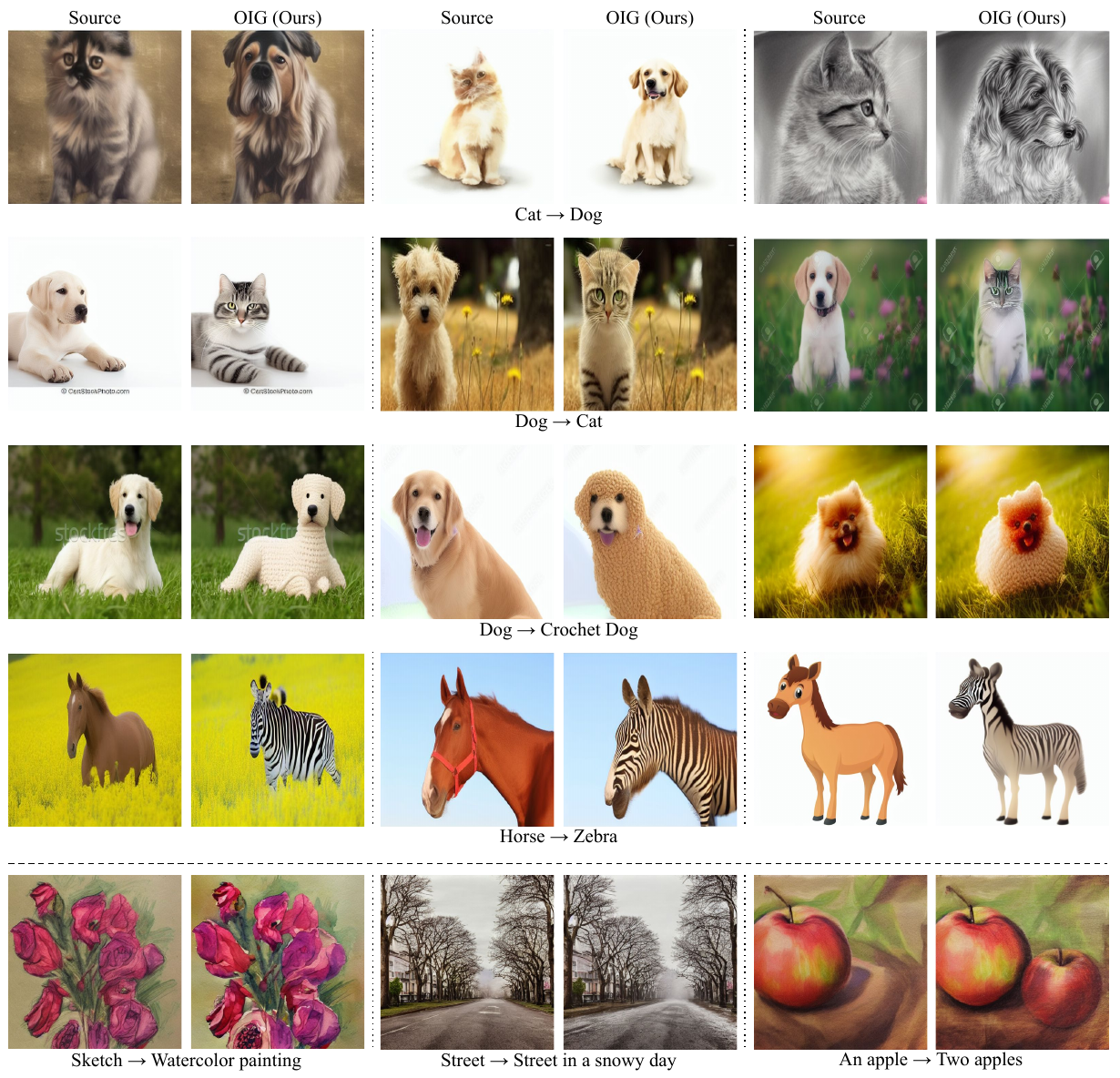}
	\caption{Qualitative results of the proposed method combined with Distilled Stable Diffusion on real images (1st - 4th row) sampled from the LAION-5B dataset~\cite{schuhmann2022laion} and synthetic images (5th row) given by the pretrained Distilled Stable Diffusion.} 
	\label{fig:qualitative_SD_Distilled}
\end{figure*}

\begin{figure*}[h!]
	\centering
	\includegraphics[width=1.0\linewidth]{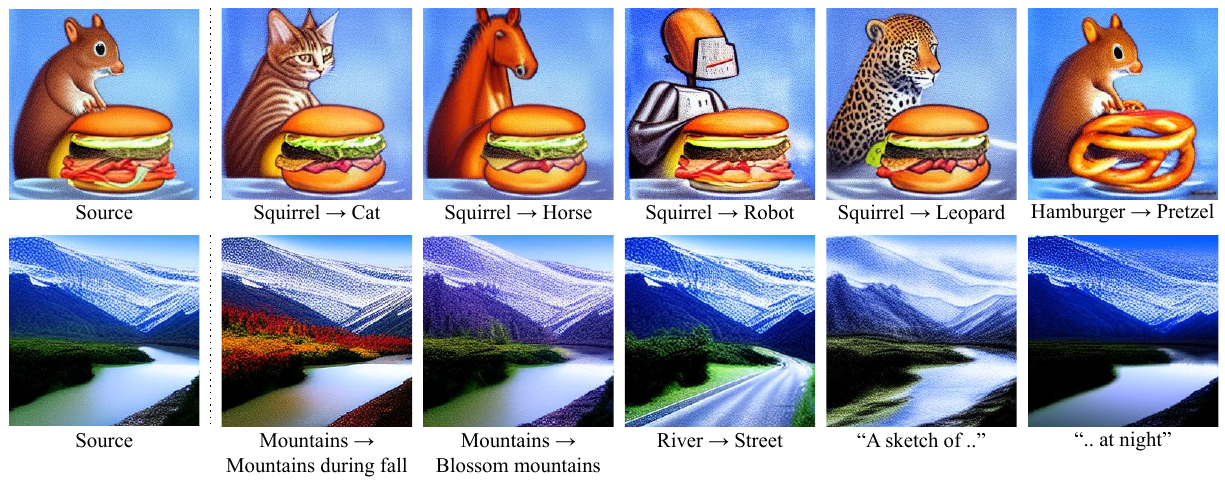}
	\caption{Qualitative results of the proposed method combined with LDM~\cite{rombach2022high} on synthetic images given by the pretrained LDM.} 
	\label{fig:qualitative_LDM}
\end{figure*}

As visualized in Figure~\ref{fig:qualitative_SD_Distilled} and~\ref{fig:qualitative_LDM}, the proposed method shows superior performance when combined with Distilled Stable Diffusion and LDM, which demonstrates that our method can generalize well.


\section{Additional qualitative results}
\label{sec:appendix_qualitative}
\begin{figure*}[hbt!]
	\centering
	\includegraphics[width=0.9\linewidth]{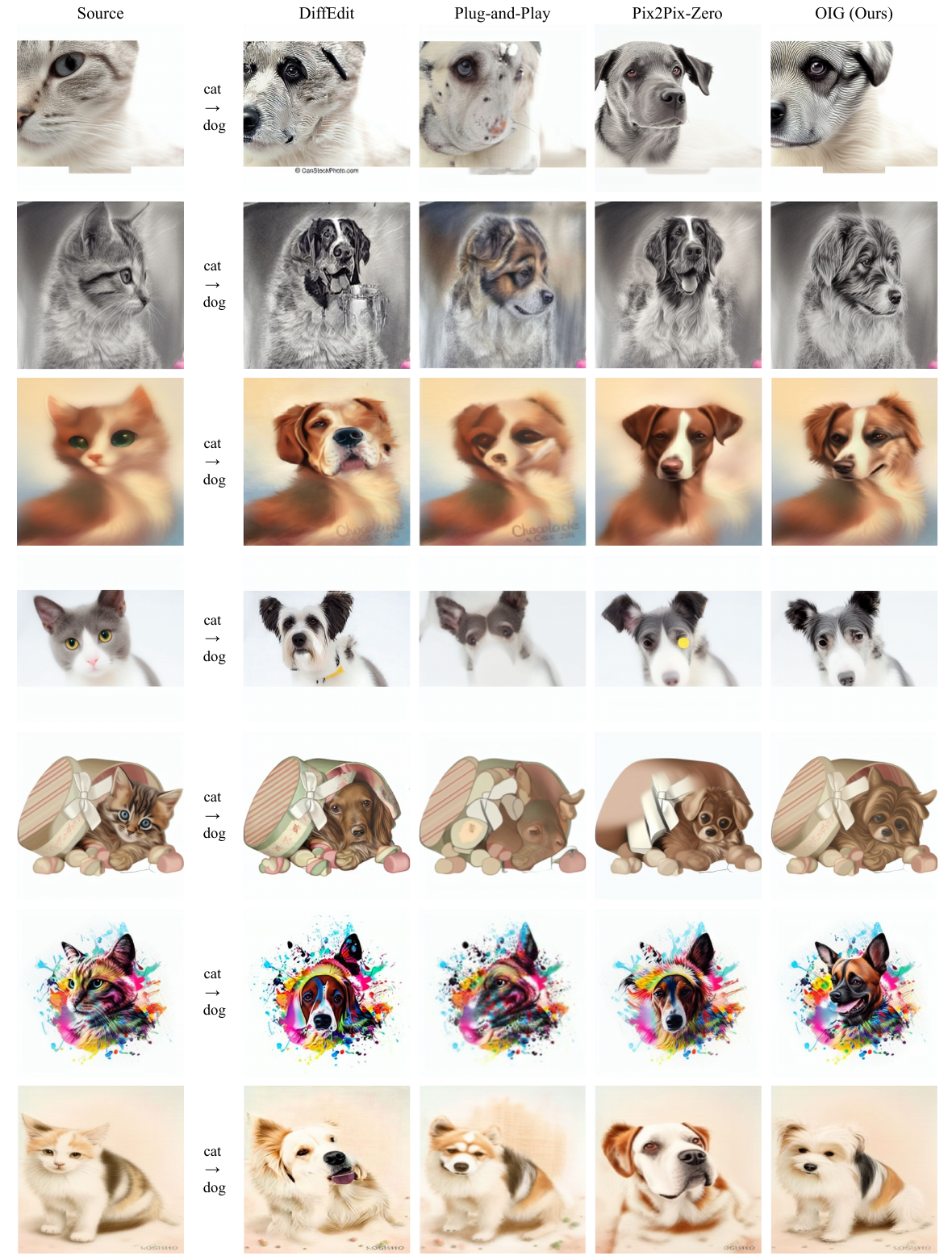}
	\caption{Additional qualitative results of the proposed method, DiffEdit~\cite{couairon2022diffedit}, Plug-and-Play~\cite{tumanyan2023plug}, and Pix2Pix-Zero~\cite{parmar2023zero} using the pretrained Stable Diffusion~\cite{rombach2022high} and real images sampled from the LAION 5B dataset~\cite{schuhmann2022laion} on the cat $\rightarrow$ dog task.} 
\label{fig:cat_to_dog_supple}
\end{figure*}

\begin{figure*}
	\centering
	\includegraphics[width=0.9\linewidth]{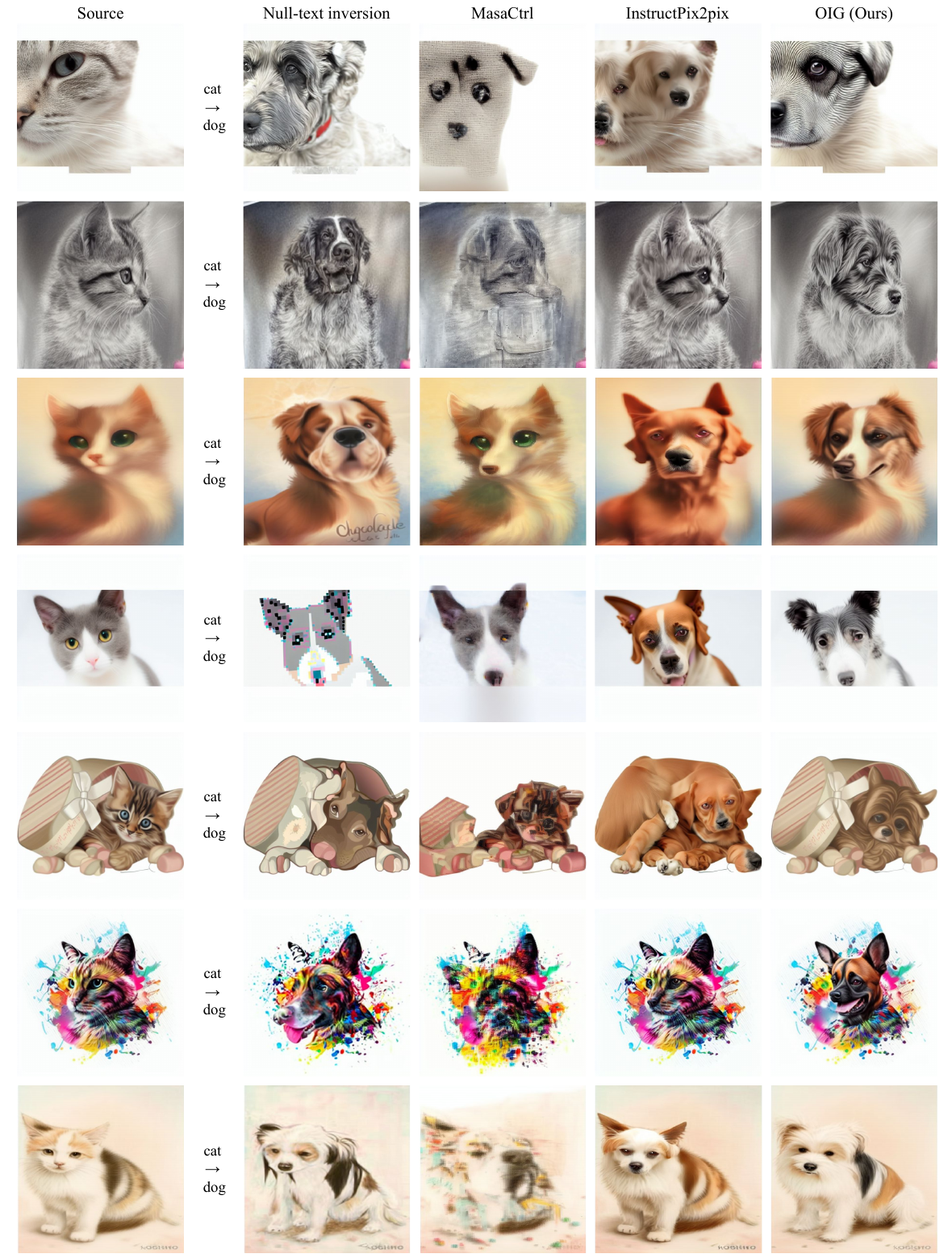}
	\caption{Additional qualitative results of the proposed method, Null-text inversion~\cite{mokady2023null}, MasaCtrl~\cite{cao2023masactrl}, and InstructPix2Pix~\cite{brooks2023instructpix2pix} using the pretrained Stable Diffusion~\cite{rombach2022high} and real images sampled from the LAION 5B dataset~\cite{schuhmann2022laion} on the cat $\rightarrow$ dog task.} 
\label{fig:cat_to_dog_supple_2}
\end{figure*}

\begin{figure*}
	\centering
	\includegraphics[width=0.9\linewidth]{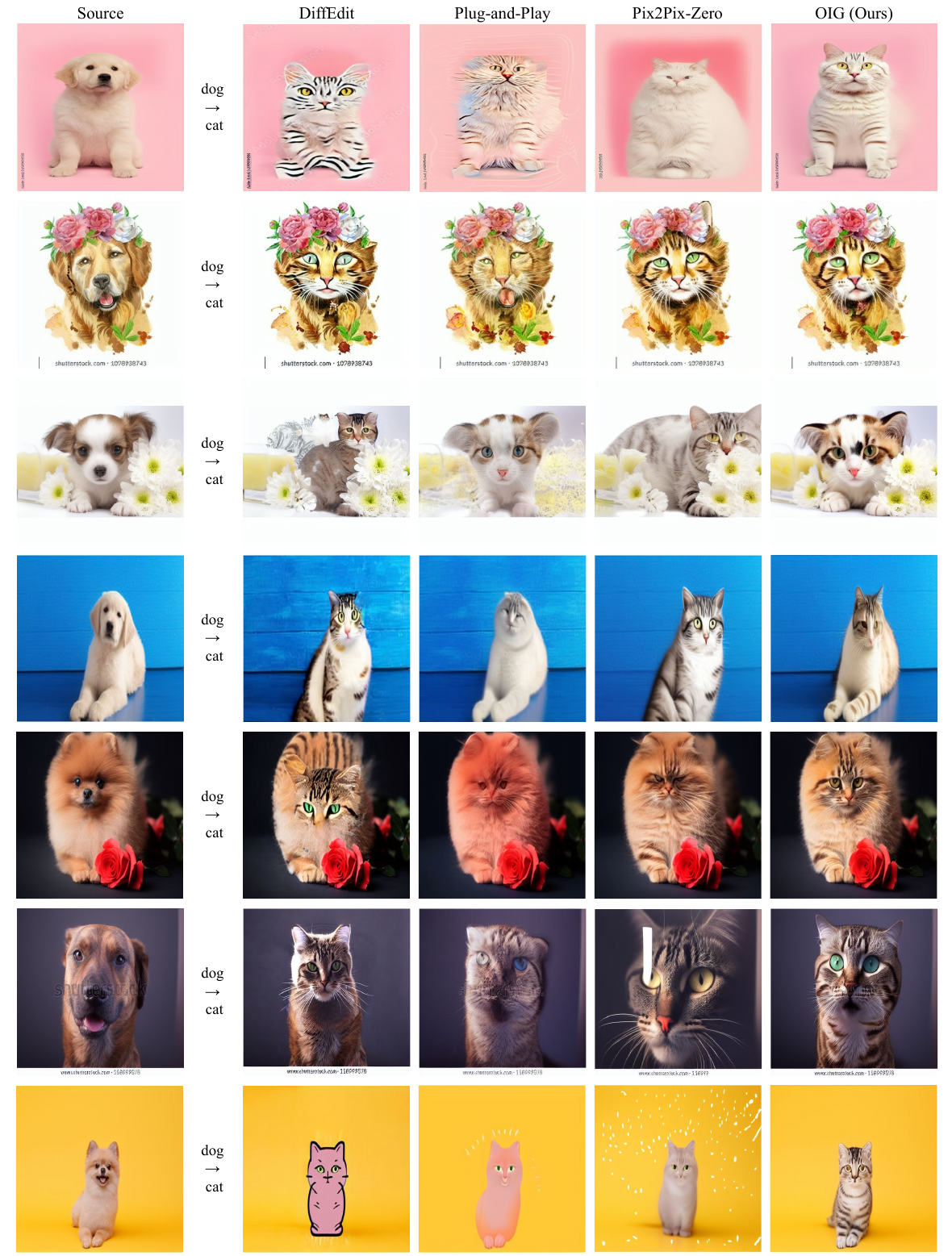}
	\caption{Additional qualitative results of the proposed method, DiffEdit~\cite{couairon2022diffedit}, Plug-and-Play~\cite{tumanyan2023plug}, and Pix2Pix-Zero~\cite{parmar2023zero} using the pretrained Stable Diffusion~\cite{rombach2022high} and real images sampled from the LAION 5B dataset~\cite{schuhmann2022laion} on the dog $\rightarrow$ cat task.} 
\label{fig:dog_to_cat_supple}
\end{figure*}

\begin{figure*}
	\centering
	\includegraphics[width=0.9\linewidth]{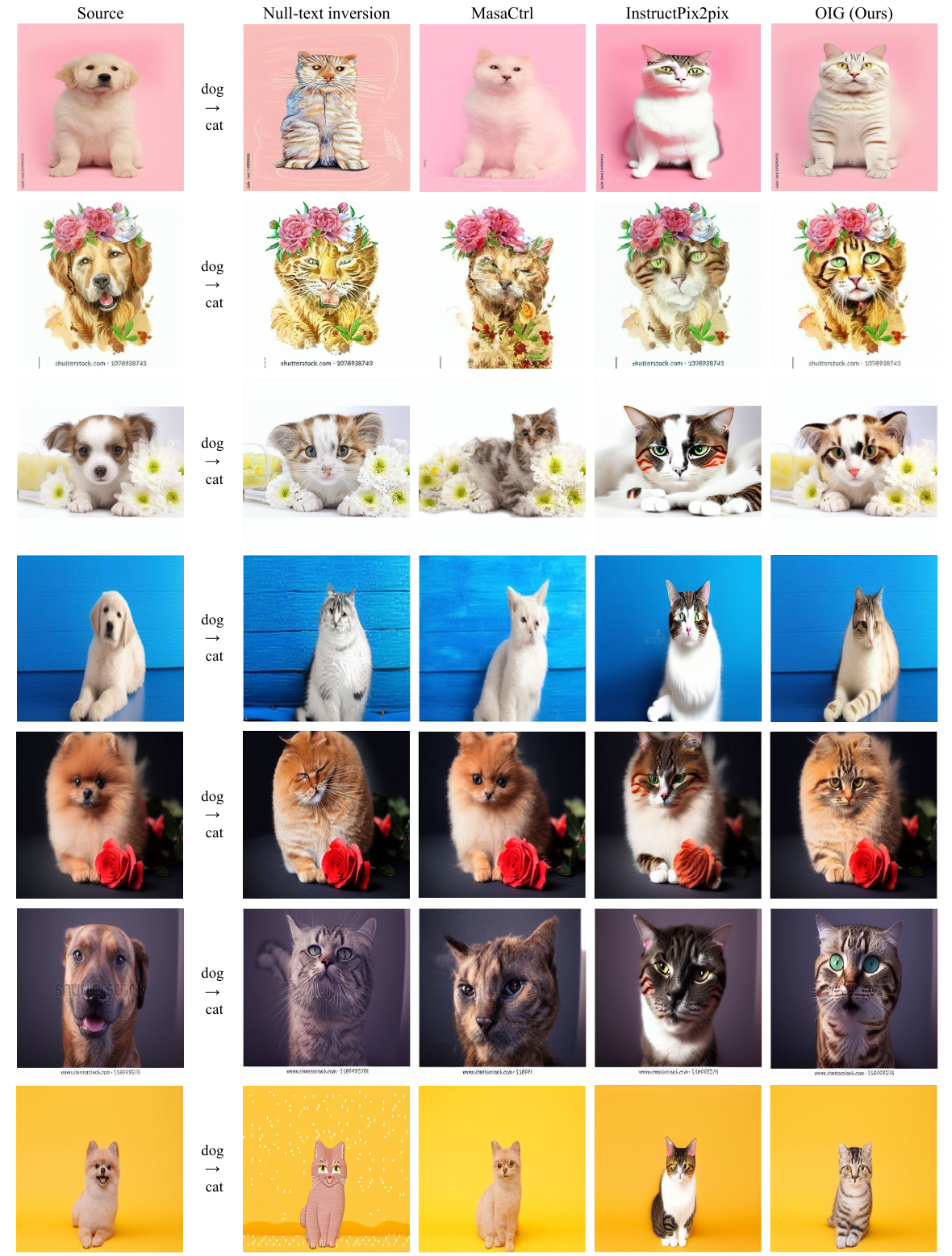}
	\caption{Additional qualitative results of the proposed method, Null-text inversion~\cite{mokady2023null}, MasaCtrl~\cite{cao2023masactrl}, and InstructPix2Pix~\cite{brooks2023instructpix2pix} using the pretrained Stable Diffusion~\cite{rombach2022high} and real images sampled from the LAION 5B dataset~\cite{schuhmann2022laion} on the dog $\rightarrow$ cat task.} 
\label{fig:dog_to_cat_supple_2}
\end{figure*}

\begin{figure*}
	\centering
	\includegraphics[width=0.9\linewidth]{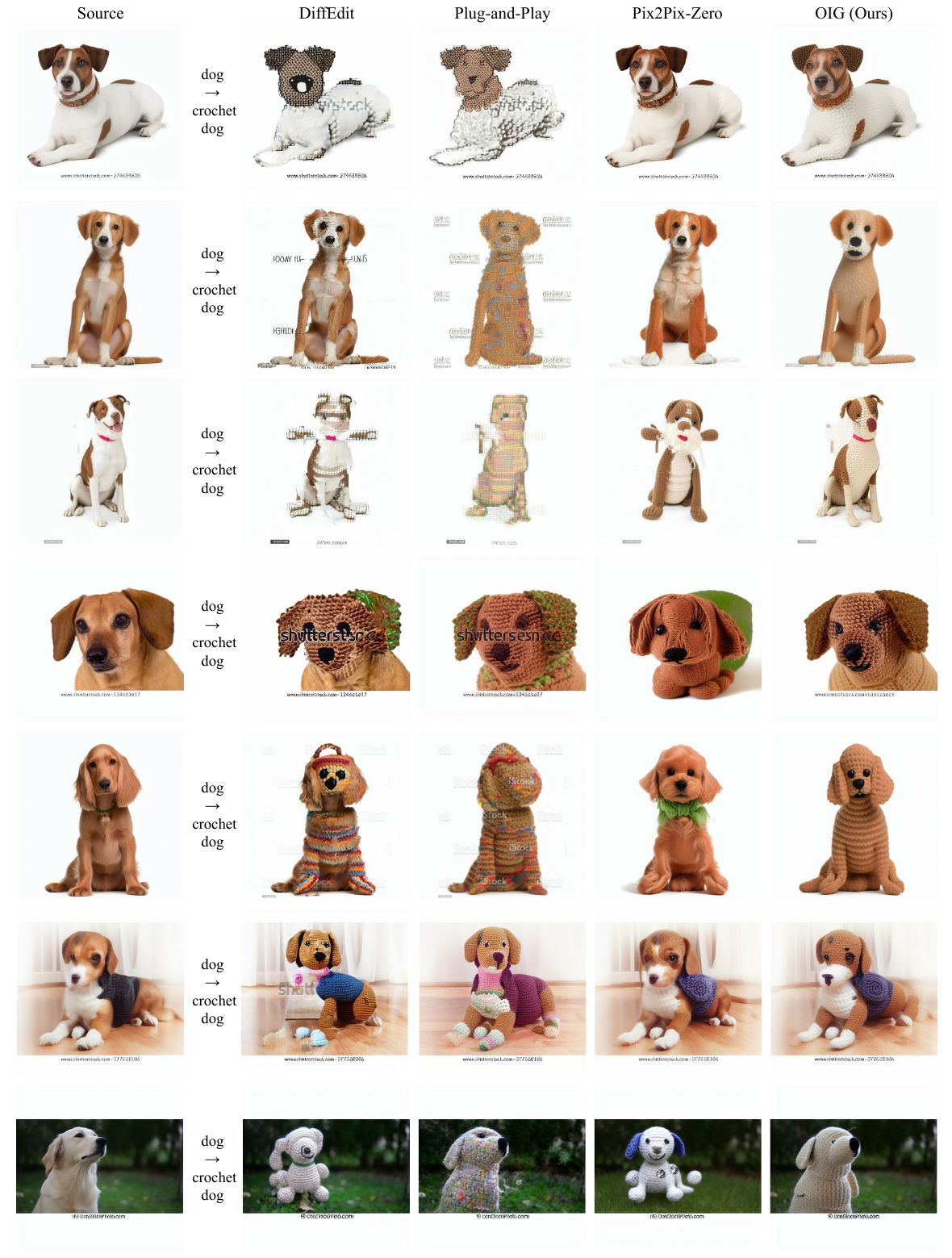}
	\caption{Additional qualitative results of the proposed method, DiffEdit~\cite{couairon2022diffedit}, Plug-and-Play~\cite{tumanyan2023plug}, and Pix2Pix-Zero~\cite{parmar2023zero} using the pretrained Stable Diffusion~\cite{rombach2022high} and real images sampled from the LAION 5B dataset~\cite{schuhmann2022laion} on the dog $\rightarrow$ crochet dog task.} 
\label{fig:dog_to_crochetdog_supple}
\end{figure*}

\begin{figure*}
	\centering
	\includegraphics[width=0.9\linewidth]{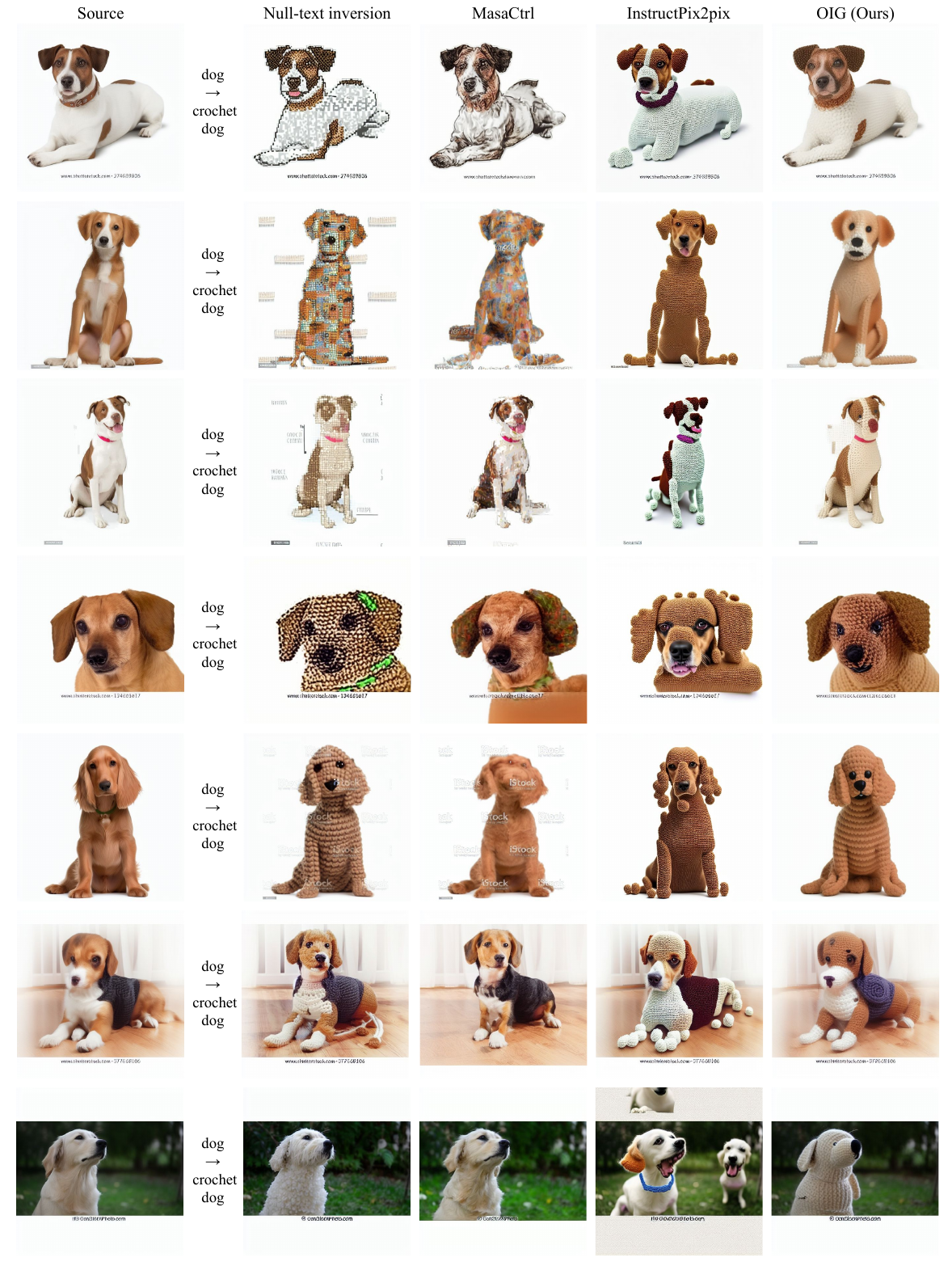}
	\caption{Additional qualitative results of the proposed method, Null-text inversion~\cite{mokady2023null}, MasaCtrl~\cite{cao2023masactrl}, and InstructPix2Pix~\cite{brooks2023instructpix2pix} using the pretrained Stable Diffusion~\cite{rombach2022high} and real images sampled from the LAION 5B dataset~\cite{schuhmann2022laion} on the dog $\rightarrow$ crochet dog task.} 
\label{fig:dog_to_crochetdog_supple_2}
\end{figure*}

\begin{figure*}
	\centering
	\includegraphics[width=0.9\linewidth]{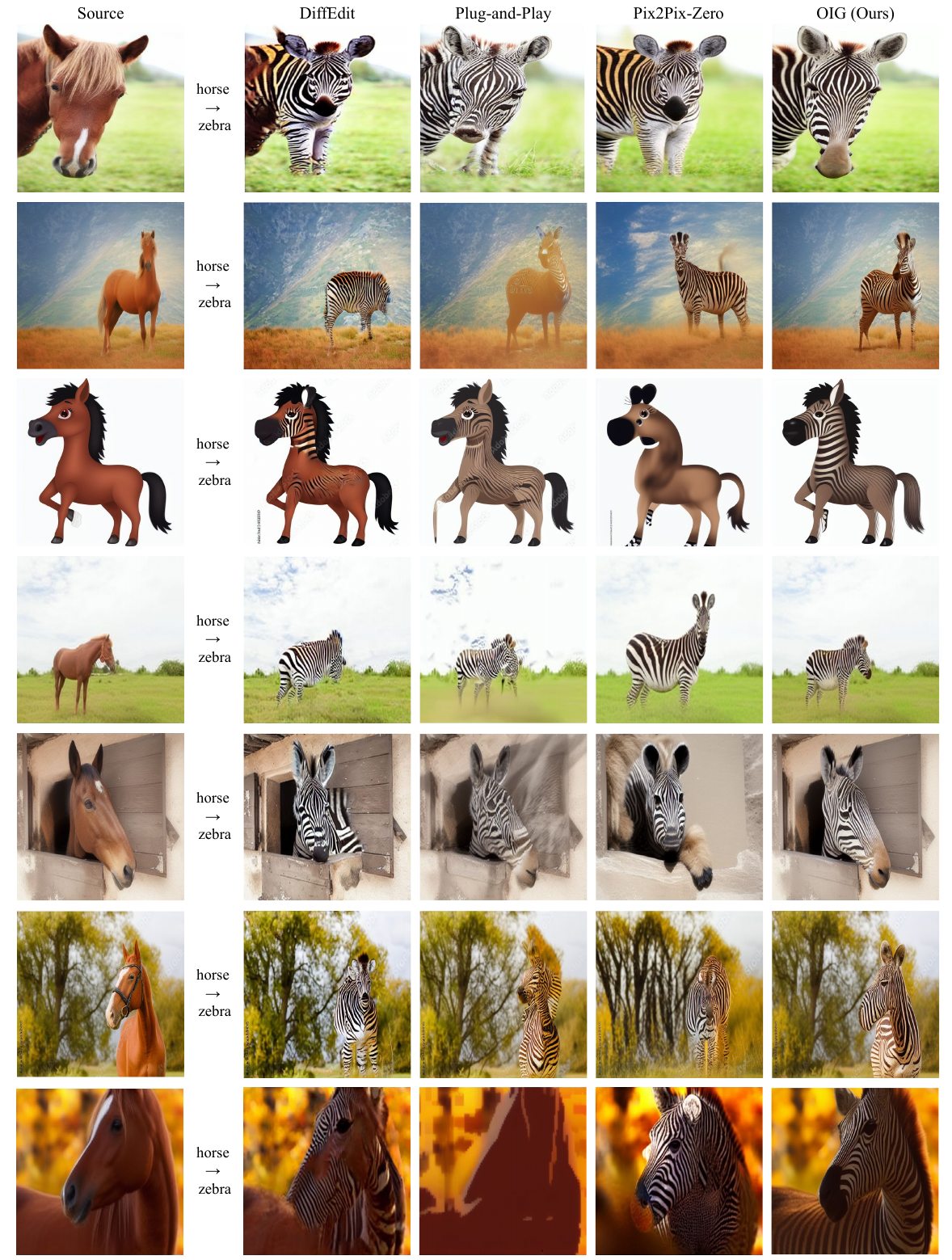}
	\caption{Additional qualitative results of the proposed method, DiffEdit~\cite{couairon2022diffedit}, Plug-and-Play~\cite{tumanyan2023plug}, and Pix2Pix-Zero~\cite{parmar2023zero} using the pretrained Stable Diffusion~\cite{rombach2022high} and real images sampled from the LAION 5B dataset~\cite{schuhmann2022laion} on the horse $\rightarrow$ zebra task.} 
\label{fig:horse_to_zebra_supple}
\end{figure*}

\begin{figure*}
	\centering
	\includegraphics[width=0.9\linewidth]{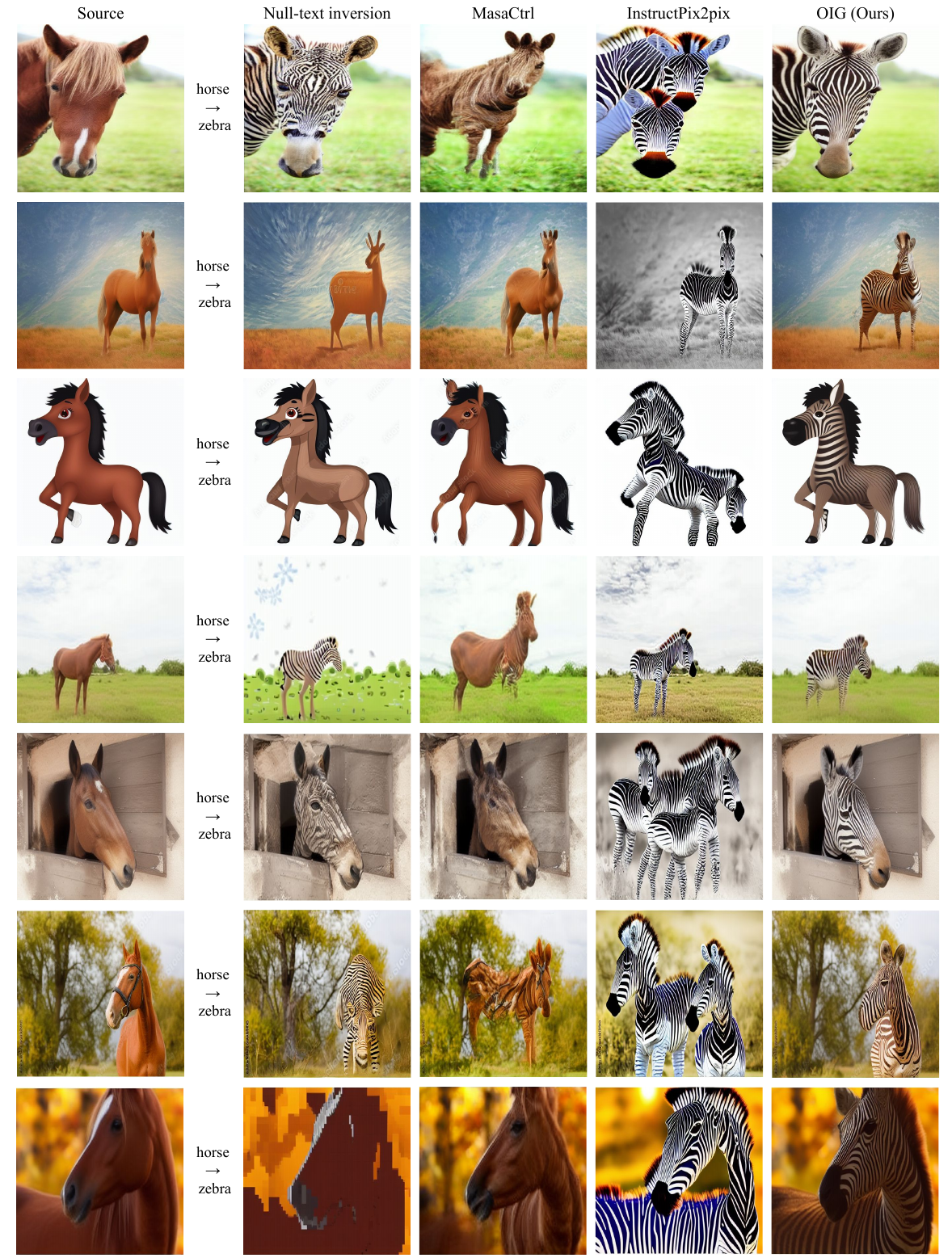}
	\caption{Additional qualitative results of the proposed method, Null-text inversion~\cite{mokady2023null}, MasaCtrl~\cite{cao2023masactrl}, and InstructPix2Pix~\cite{brooks2023instructpix2pix} using the pretrained Stable Diffusion~\cite{rombach2022high} and real images sampled from the LAION 5B dataset~\cite{schuhmann2022laion} on the horse $\rightarrow$ zebra task.} 
\label{fig:horse_to_zebra_supple_2}
\end{figure*}

\begin{figure*}
	\centering
	\includegraphics[width=0.9\linewidth]{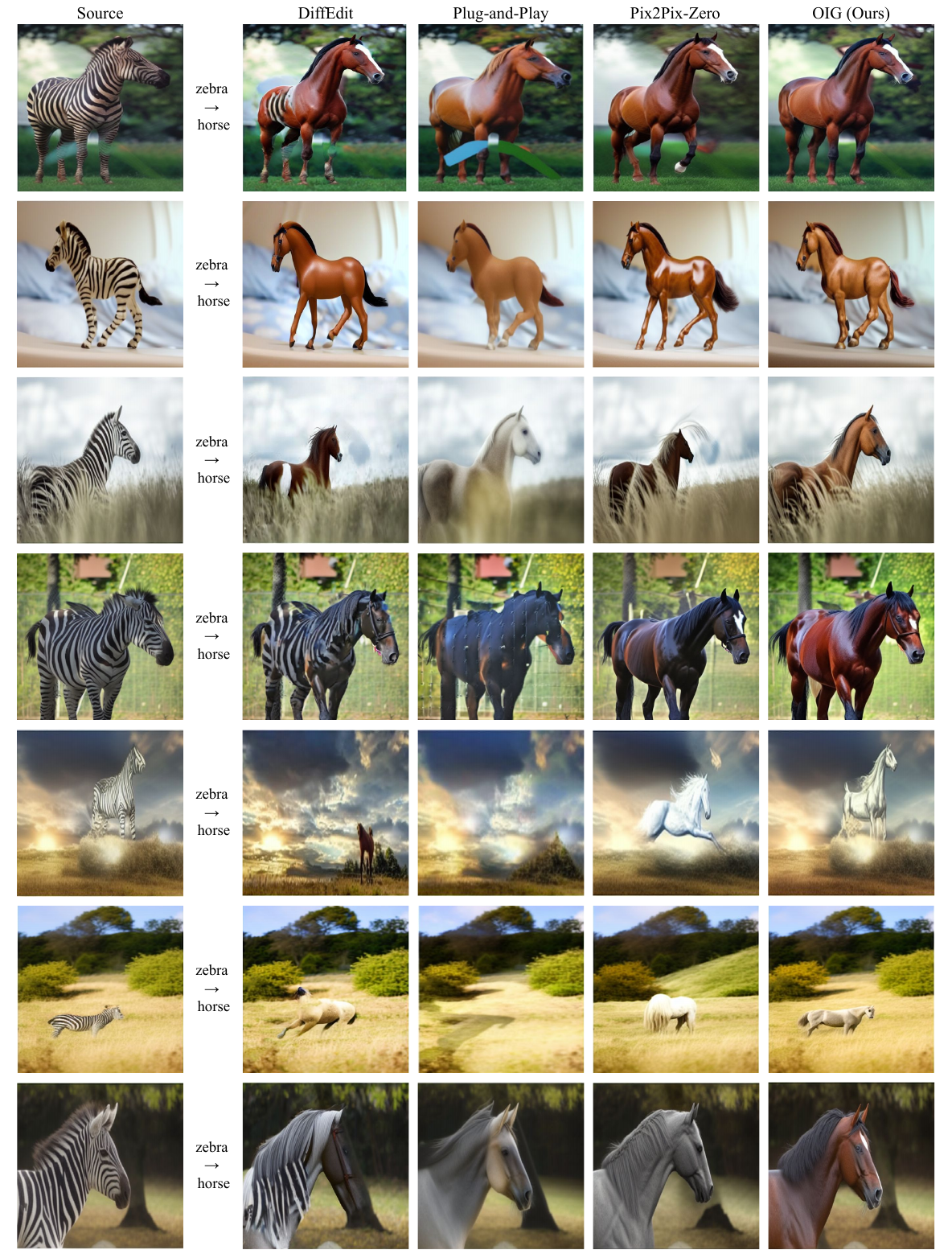}
	\caption{Additional qualitative results of the proposed method, DiffEdit~\cite{couairon2022diffedit}, Plug-and-Play~\cite{tumanyan2023plug}, and Pix2Pix-Zero~\cite{parmar2023zero} using the pretrained Stable Diffusion~\cite{rombach2022high} and real images sampled from the LAION 5B dataset~\cite{schuhmann2022laion} on the zebra $\rightarrow$ horse task.} 
\label{fig:zebra_to_horse_supple}
\end{figure*}

\begin{figure*}
	\centering
	\includegraphics[width=0.9\linewidth]{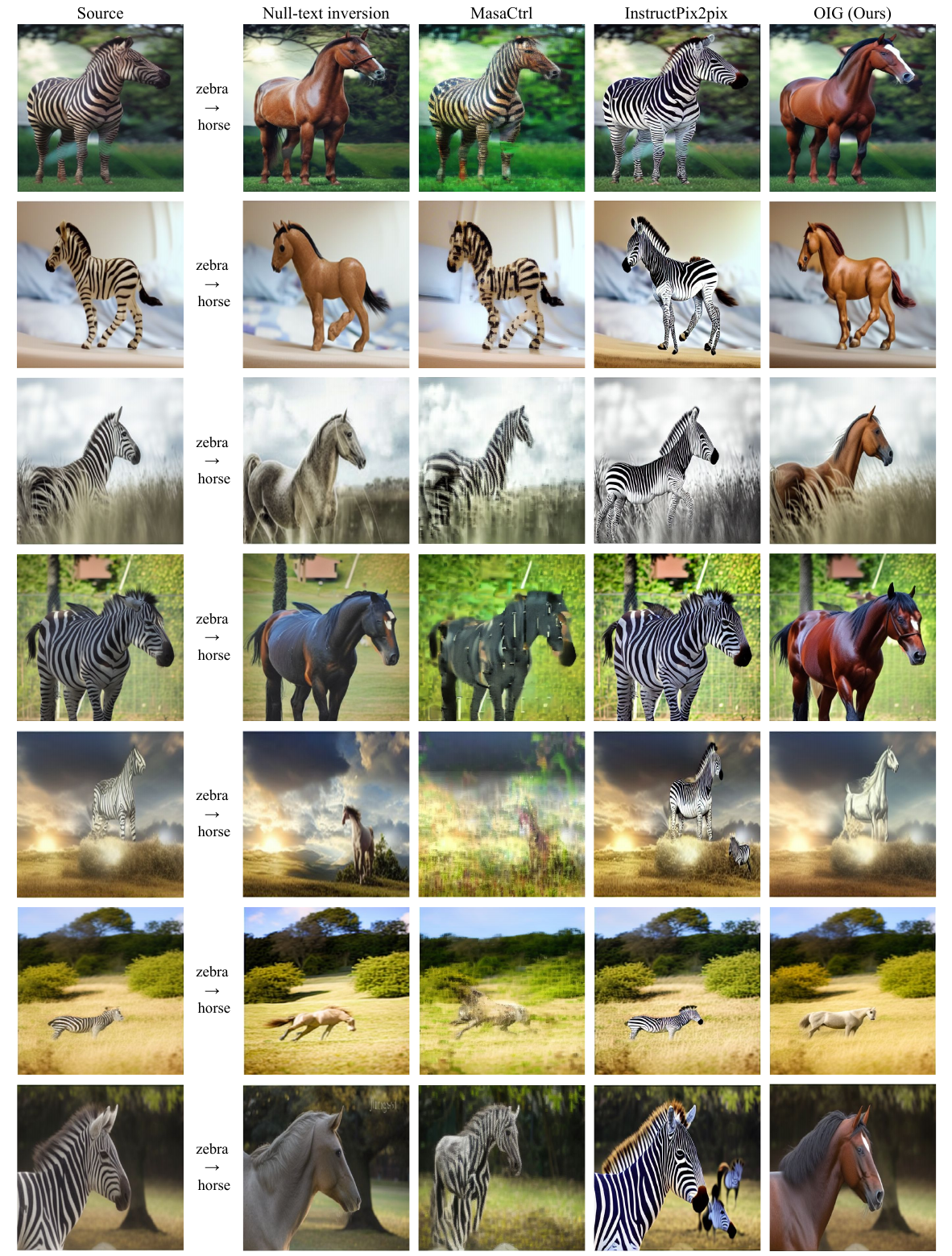}
	\caption{Additional qualitative results of the proposed method, Null-text inversion~\cite{mokady2023null}, MasaCtrl~\cite{cao2023masactrl}, and InstructPix2Pix~\cite{brooks2023instructpix2pix} using the pretrained Stable Diffusion~\cite{rombach2022high} and real images sampled from the LAION 5B dataset~\cite{schuhmann2022laion} on the zebra $\rightarrow$ horse task.} 
\label{fig:zebra_to_horse_supple_2}
\end{figure*}

\begin{figure*}
	\centering
	\includegraphics[width=0.9\linewidth]{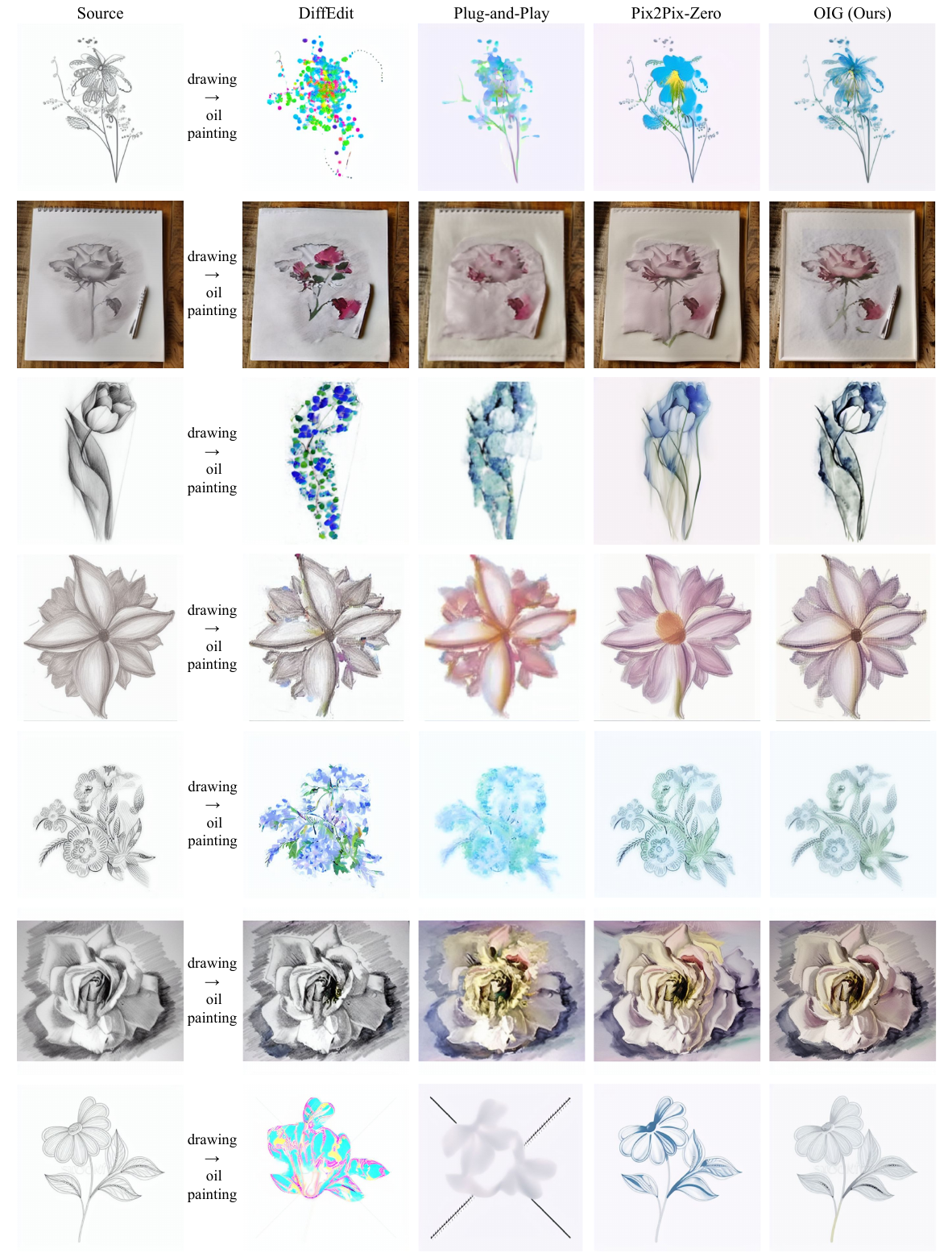}
	\caption{Additional qualitative results of the proposed method, DiffEdit~\cite{couairon2022diffedit}, Plug-and-Play~\cite{tumanyan2023plug}, and Pix2Pix-Zero~\cite{parmar2023zero} using the pretrained Stable Diffusion~\cite{rombach2022high} and real images sampled from the LAION 5B dataset~\cite{schuhmann2022laion} on the drawing $\rightarrow$ oil painting task.} 
\label{fig:drawing_to_oilpainting_supple}
\end{figure*}

\begin{figure*}
	\centering
	\includegraphics[width=0.9\linewidth]{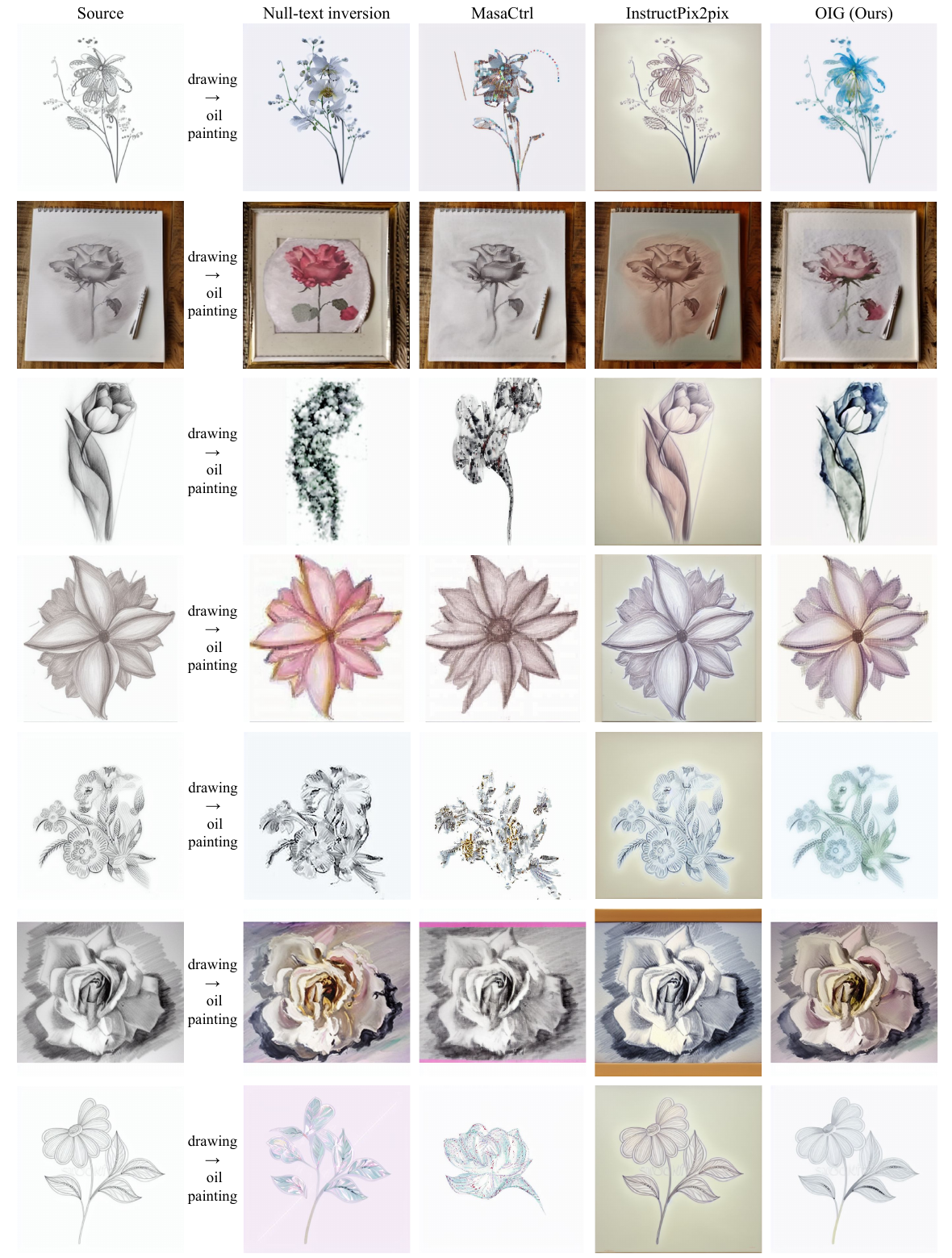}
	\caption{Additional qualitative results of the proposed method, Null-text inversion~\cite{mokady2023null}, MasaCtrl~\cite{cao2023masactrl}, and InstructPix2Pix~\cite{brooks2023instructpix2pix} using the pretrained Stable Diffusion~\cite{rombach2022high} and real images sampled from the LAION 5B dataset~\cite{schuhmann2022laion} on the drawing $\rightarrow$ oil painting task.} 
\label{fig:drawing_to_oilpainting_supple_2}
\end{figure*}

We present additional qualitative results of OIG in Figure~\ref{fig:cat_to_dog_supple},~\ref{fig:cat_to_dog_supple_2},~\ref{fig:dog_to_cat_supple},~\ref{fig:dog_to_cat_supple_2},~\ref{fig:dog_to_crochetdog_supple},~\ref{fig:dog_to_crochetdog_supple_2},~\ref{fig:horse_to_zebra_supple},~\ref{fig:horse_to_zebra_supple_2},~\ref{fig:zebra_to_horse_supple},~\ref{fig:zebra_to_horse_supple_2},~\ref{fig:drawing_to_oilpainting_supple}, and~\ref{fig:drawing_to_oilpainting_supple_2} to compare with state-of-the-art methods~\cite{couairon2022diffedit, tumanyan2023plug, parmar2023zero, mokady2023null, cao2023masactrl, brooks2023instructpix2pix} on real images sampled from the LAION-5B dataset~\cite{schuhmann2022laion} using the pretrained Stable Diffusion~\cite{rombach2022high}. 
As visualized in the figures, OIG achieves outstanding results while the previous methods often fail to preserve the structure or background of the source images.

\begin{figure*}
	\centering
	\includegraphics[width=0.75\linewidth]{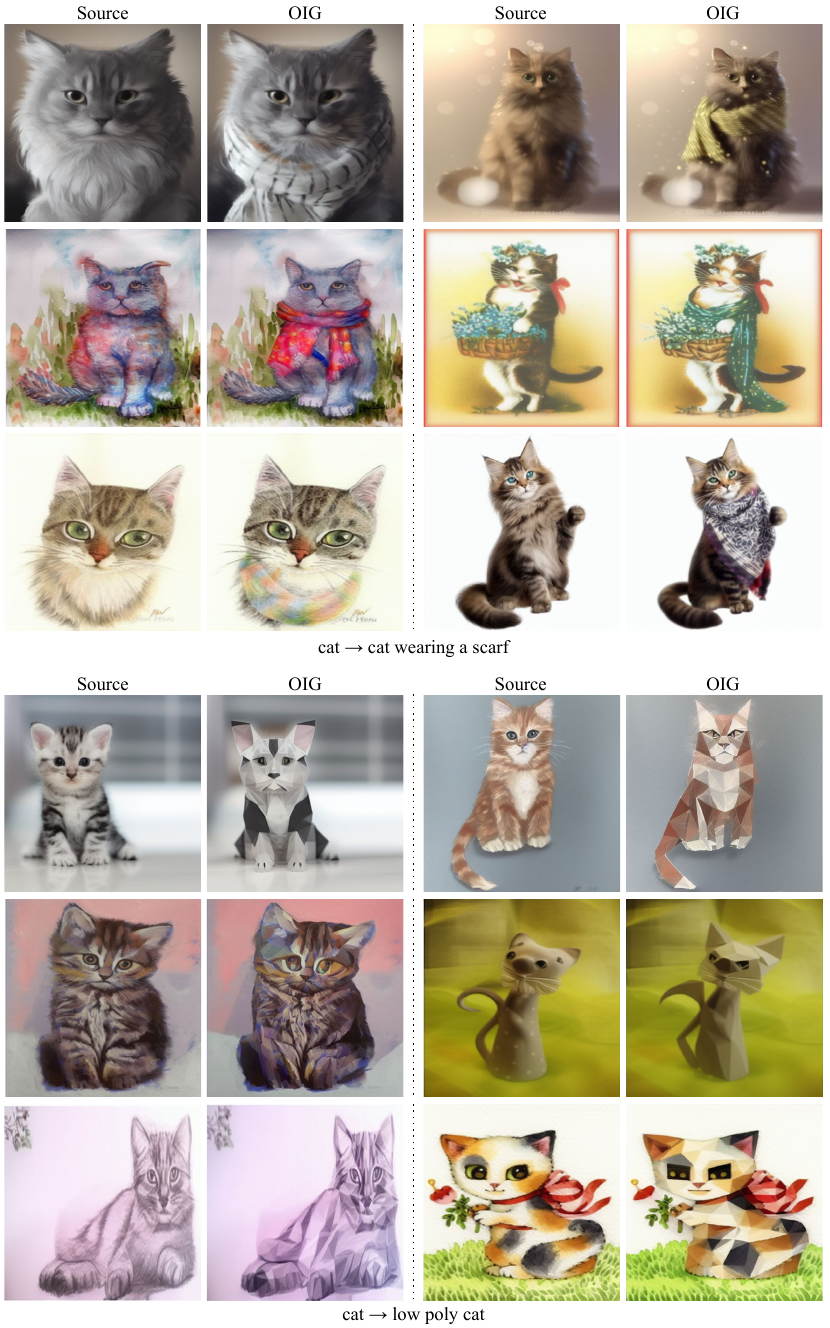}
	\caption{Qualitative results of the proposed method on real images sampled from the LAION 5B dataset~\cite{schuhmann2022laion} on the cat $\rightarrow$ cat wearing a scarf and cat $\rightarrow$ low poly cat task.} 
\label{fig:real_laion5b_additional_1}
\end{figure*}

\begin{figure*}
	\centering
	\includegraphics[width=0.75\linewidth]{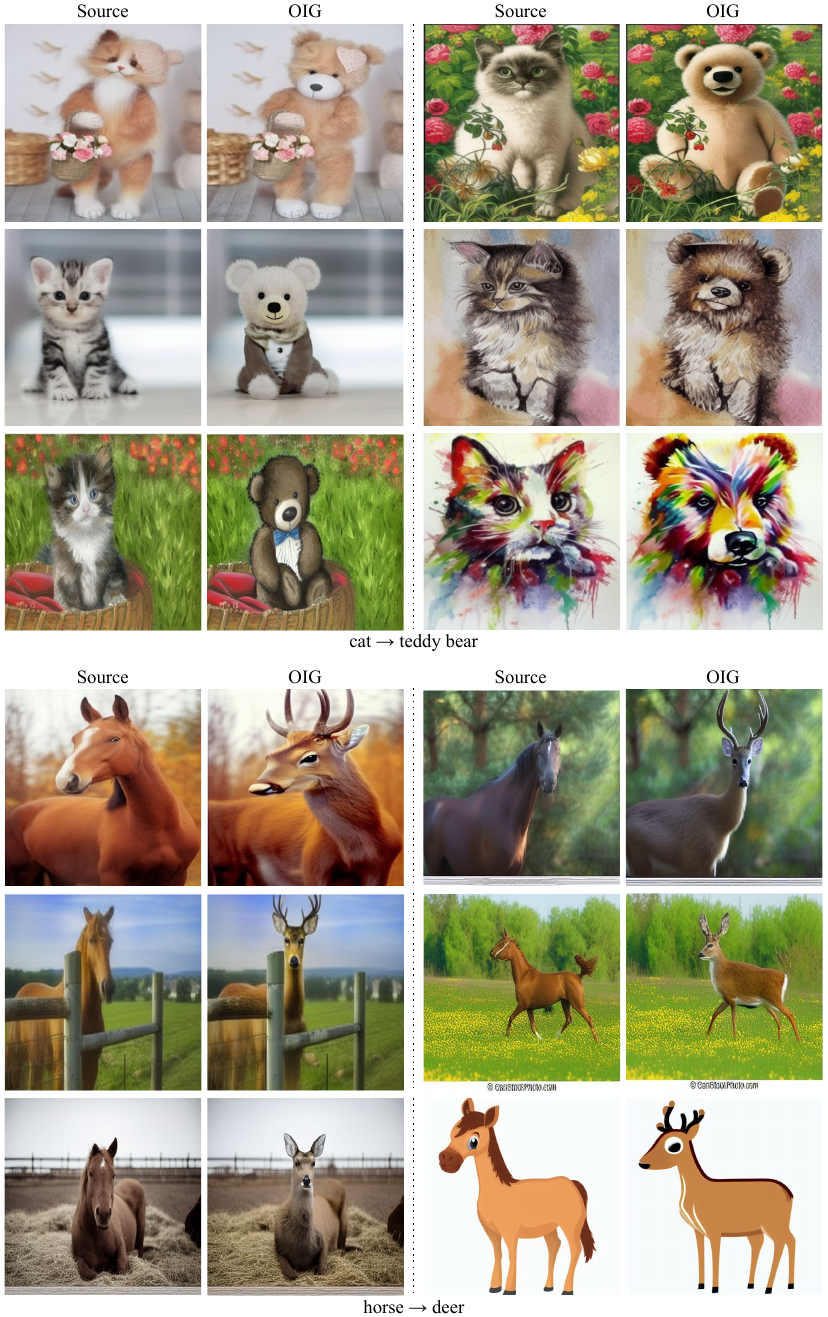}
	\caption{Qualitative results of the proposed method on real images sampled from the LAION 5B dataset~\cite{schuhmann2022laion} on the cat $\rightarrow$ teddy bear and horse $\rightarrow$ deer task.} 
\label{fig:real_laion5b_additional_2}
\end{figure*}

\begin{figure*}
	\centering
	\includegraphics[width=0.85\linewidth]{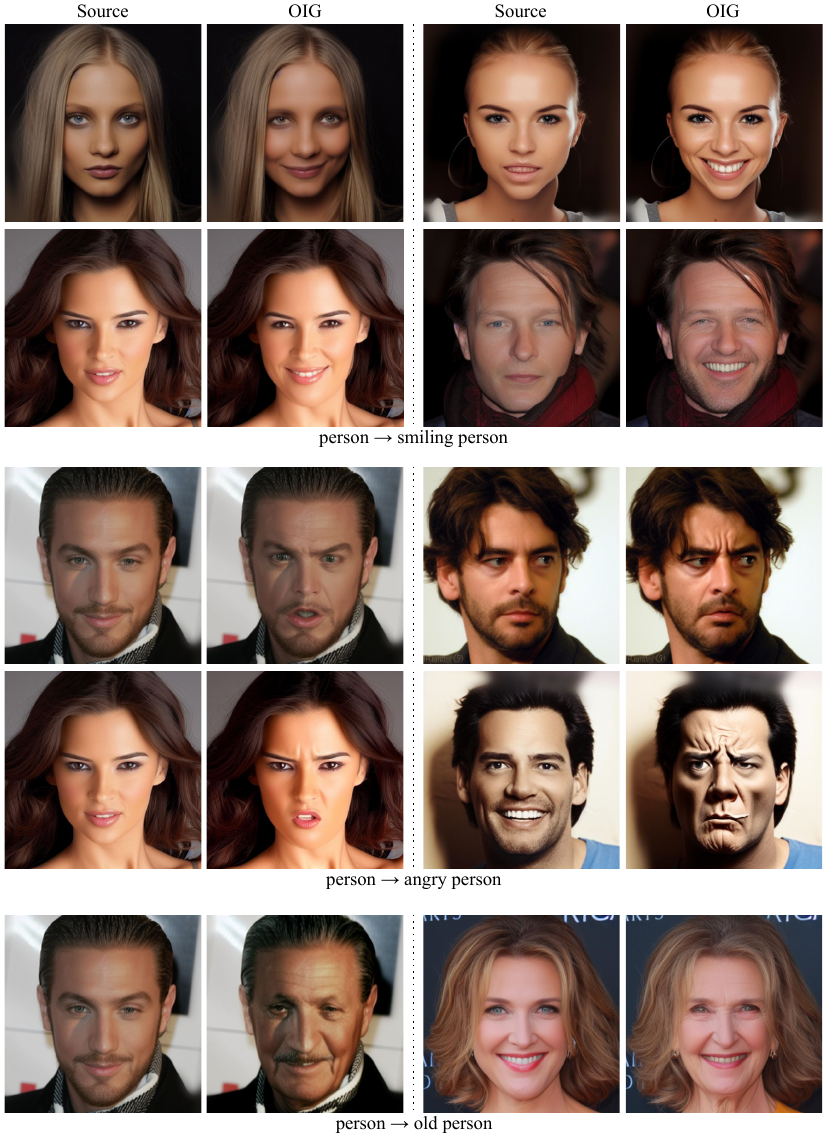}
	\caption{Qualitative results of the proposed method on real images sampled from the CelebA-HQ dataset~\cite{karras2017progressive} using the pretrained Stable Diffusion~\cite{rombach2022high}.}
\label{fig:celeba_hq_1}
\end{figure*}

\begin{figure*}
	\centering
	\includegraphics[width=0.85\linewidth]{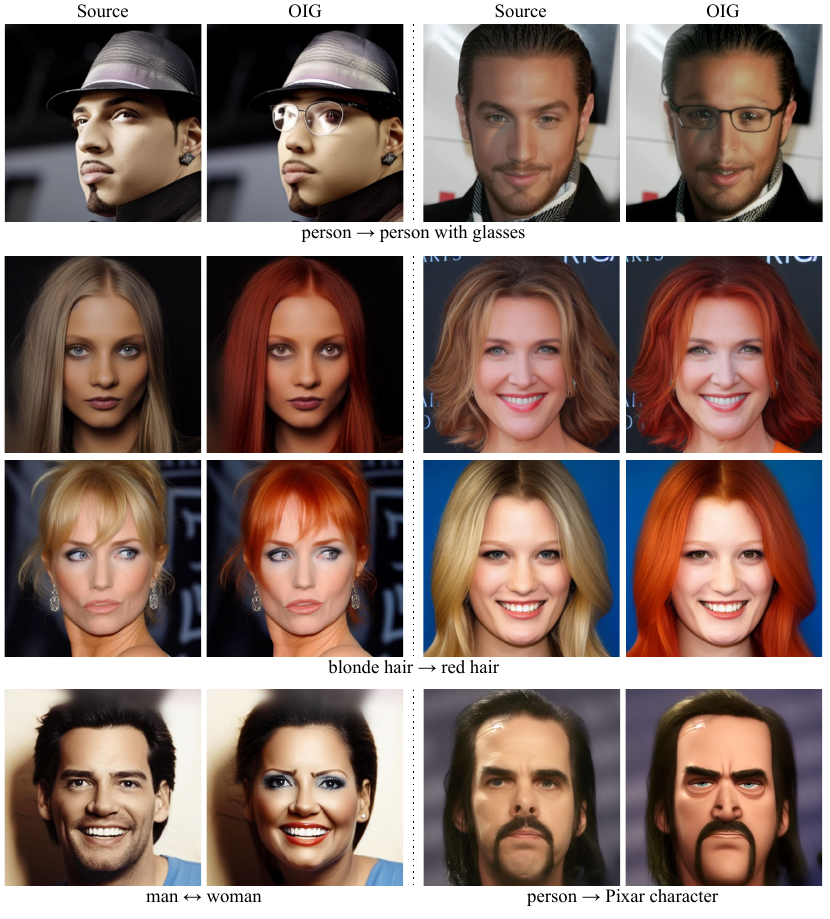}
	\caption{Qualitative results of the proposed method on real images sampled from the CelebA-HQ dataset~\cite{karras2017progressive} using the pretrained Stable Diffusion~\cite{rombach2022high}.}
\label{fig:celeba_hq_2}
\end{figure*}

\begin{figure*}
	\centering
	\includegraphics[width=0.75\linewidth]{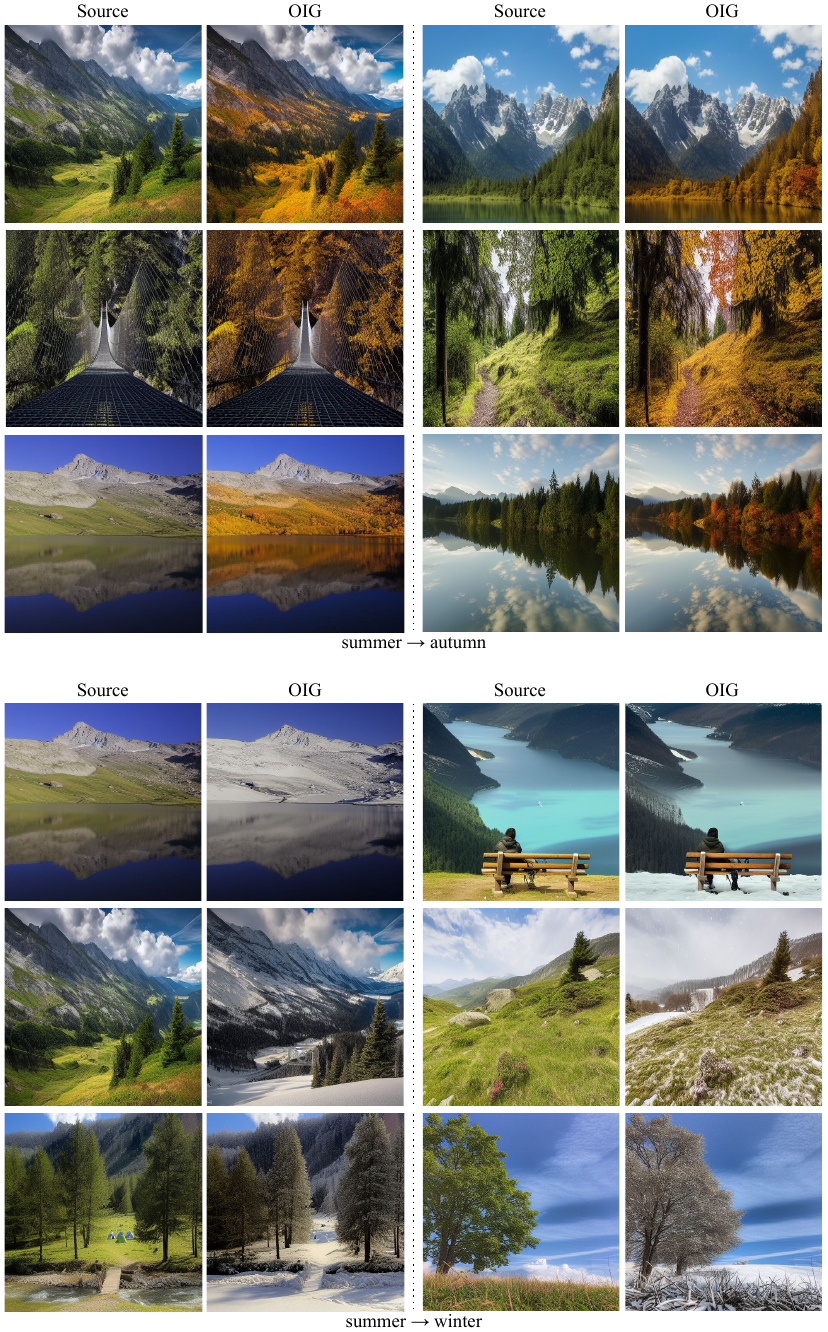}
	\caption{Qualitative results of the proposed method with real images sampled from the Seasons dataset~\cite{anoosheh2018combogan} for style transfer tasks using the pretrained Stable Diffusion~\cite{rombach2022high}.}
\label{fig:real_seasons_style_transfer_1}
\end{figure*}

\begin{figure*}
	\centering
	\includegraphics[width=0.75\linewidth]{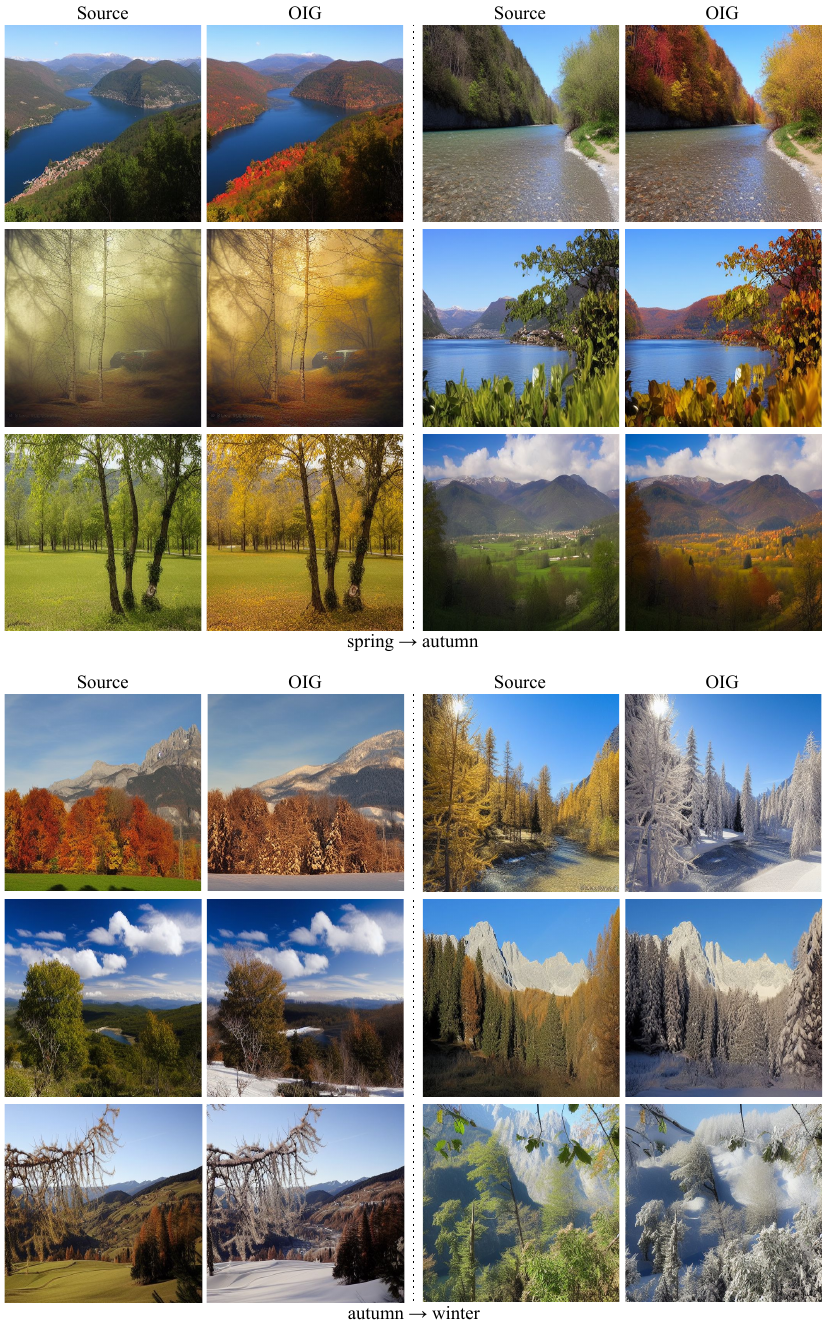}
	\caption{Qualitative results of the proposed method with real images sampled from the Seasons dataset~\cite{anoosheh2018combogan} on style transfer tasks using the pretrained Stable Diffusion~\cite{rombach2022high}.}
\label{fig:real_seasons_style_transfer_2}
\end{figure*}

\begin{figure*}
	\centering
	\includegraphics[width=0.8\linewidth]{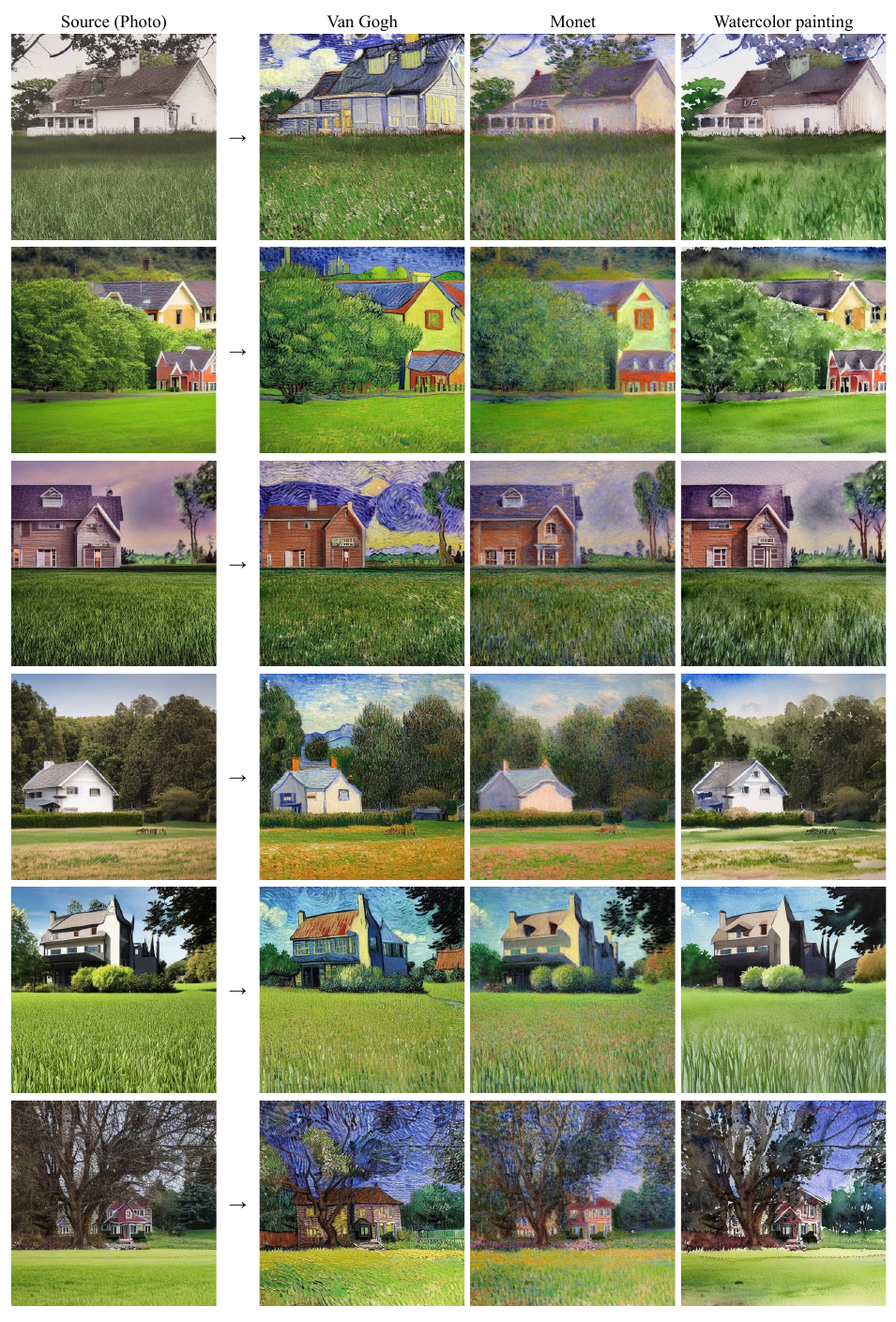}
	\caption{Qualitative results of the proposed method with synthetic images given by the pretrained Stable Diffusion~\cite{rombach2022high} on style transfer tasks.} 
\label{fig:synth_1}
\end{figure*}

\begin{figure*}
	\centering
	\includegraphics[width=0.9\linewidth]{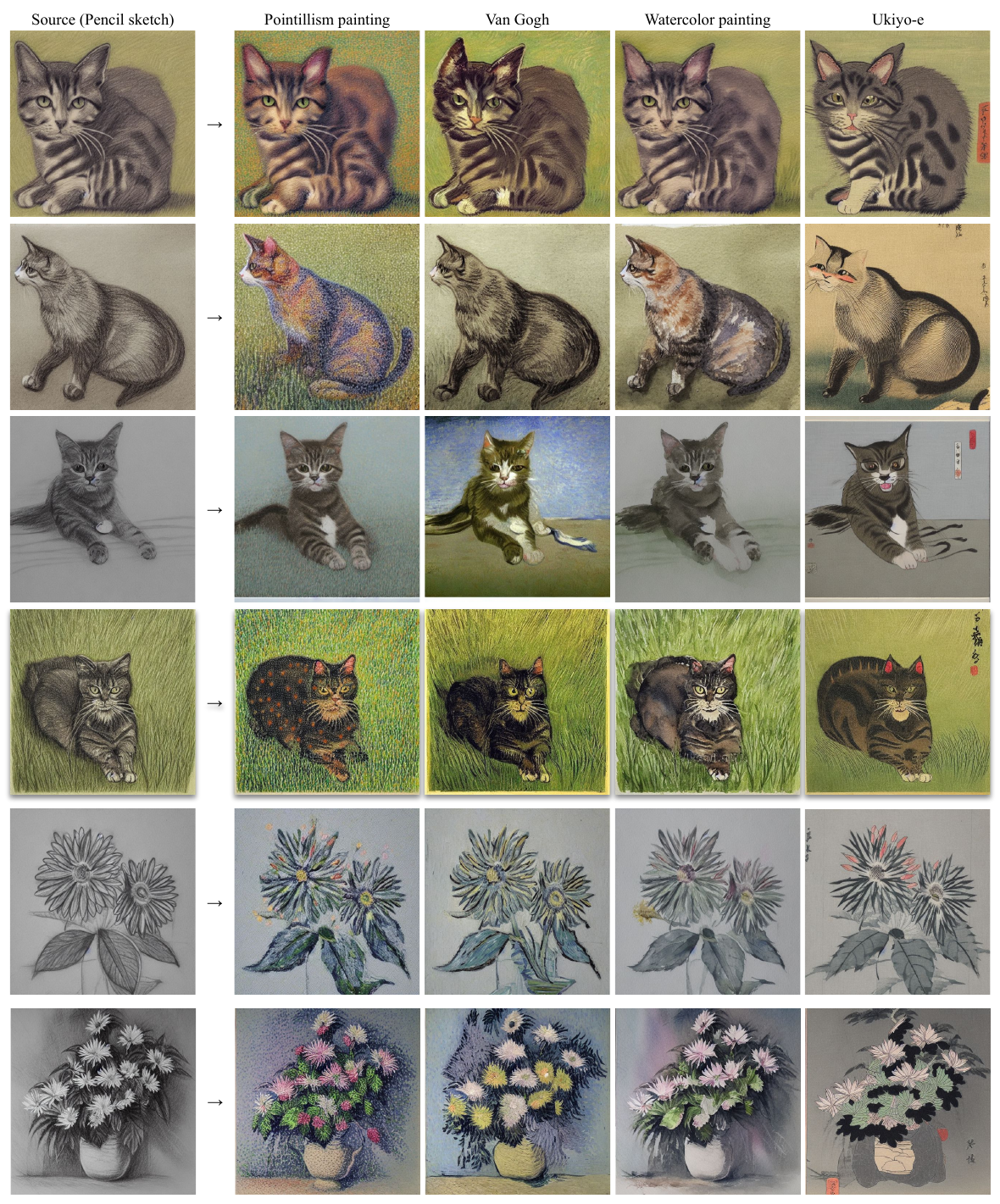}
	\caption{Qualitative results of the proposed method with synthetic images given by the pretrained Stable Diffusion~\cite{rombach2022high} on style transfer tasks.}
\label{fig:synth_2}
\end{figure*}

\begin{figure*}
	\centering
	\includegraphics[width=0.95\linewidth]{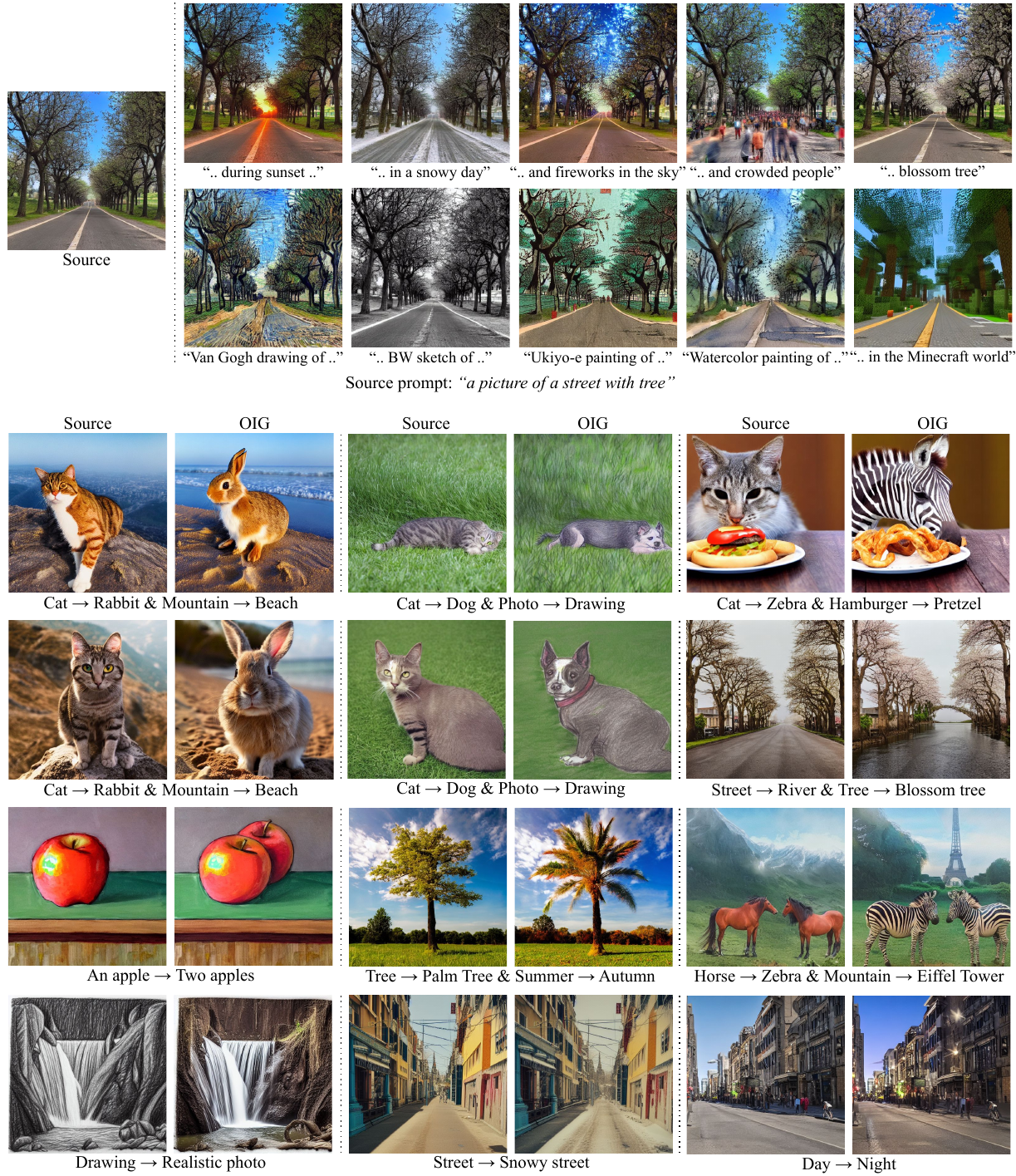}
	\caption{Qualitative results of the proposed method with synthetic images given by the pretrained Stable Diffusion~\cite{rombach2022high}.} 
\label{fig:synth_3}
\end{figure*}

In order to demonstrate the generalizability of the proposed method, we perform additional experiments using LAION-5B, CelebA-HQ~\cite{karras2017progressive}, and Seasons~\cite{anoosheh2018combogan} datasets.
We emphasize that we focus on evaluating image-to-image translation tasks both on object-centric tasks such as cat $\rightarrow$ cat wearing a scarf task and style transfer tasks like summer $\rightarrow$ winter task.
As visualized in Figure~\ref{fig:real_laion5b_additional_1}, ~\ref{fig:real_laion5b_additional_2}, ~\ref{fig:celeba_hq_1}, ~\ref{fig:celeba_hq_2}, ~\ref{fig:real_seasons_style_transfer_1}, and~\ref{fig:real_seasons_style_transfer_2}, the proposed method demonstrates superior text-driven image manipulation performance on real images for various tasks.
Figure~\ref{fig:synth_1}, ~\ref{fig:synth_2} and~\ref{fig:synth_3} also verify the effectiveness of OIG on the synthesized images given by the pretrained Stable Diffusion.

\begin{figure*}
	\centering
	\includegraphics[width=1.0\linewidth]{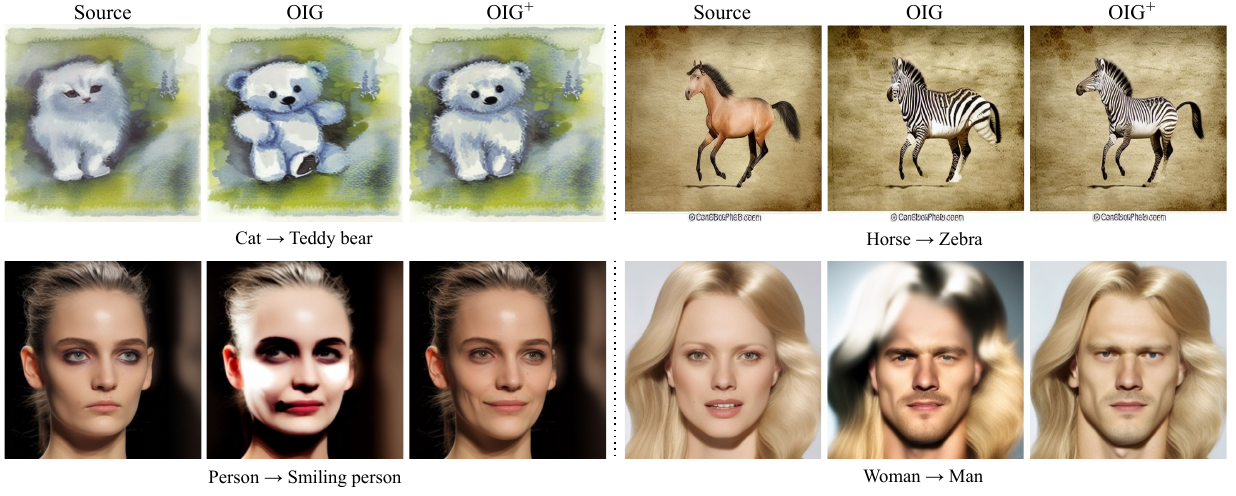}
	\caption{Failure cases of the proposed method on real images sampled from the LAION 5B dataset~\cite{schuhmann2022laion} (1st row) and CelebA-HQ dataset~\cite{karras2017progressive} (2nd row). 
	Provided failure samples can be addressed by utilizing OIG$^{+}$.}
	\label{fig:failure}
\end{figure*}

\section{Limitations}
\label{sec:appendix_limitations}
We visualize the failure cases of our method in Figure~\ref{fig:failure}. 
These failure cases can be addressed by using OIG$^+$, which combines representation guidance and coherence guidance.
OIG$^+$ effectively removes the artifacts and resolves inconsistencies in the target image, thereby improving the editing performance of the proposed method.

In addition, since the DDIM inference process sometimes does not completely reconstruct the original image, our method can struggle to preserve the information about the source image and result in suboptimal image-to-image translation results.
Furthermore, the performance of the proposed method is reliant on pretrained text-to-image diffusion models, which may limit its ability to generate target images for complex tasks effectively.

\section{Social impacts}
\label{sec:appendix_social}
The proposed method may synthesize undesirable or inappropriate images depending on the pretrained text-to-image
generation model~\cite{rombach2022high}.
For example, the incompleteness of the pretrained diffusion model can lead to the generation of images that violate ethical regulations.

\clearpage


\end{document}